%% file: hb_v3.tex
\documentclass[pmlr,twocolumn,10pt]{jmlr} 

\mlhtrack{proceedings}

\newif\iffinal
\finalfalse  

\iffinal
    \ifmlhneedspmlr
      \jmlrvolume{XXX}
      \jmlryear{2026}
    \fi
    \ifmlhfindings \jmlrproceedings{}{ML4H 2026 - Findings Track}\fi
    \ifmlhdemo     \jmlrproceedings{}{ML4H 2026 - Demo Track}\fi
    \jmlrworkshop{Machine Learning for Health (ML4H) 2026}
\else
    \jmlrproceedings{}{Submitted to ML4H 2026: \mlhtrackname}
    \jmlrworkshop{Machine Learning for Health (ML4H) 2026}
\fi

\usepackage{booktabs}
\usepackage{placeins}
\usepackage{algorithm}
\usepackage{xcolor}
\usepackage{tcolorbox}
\usepackage{amsmath,amssymb}
\usepackage{siunitx}

\usepackage{multirow}
\usepackage{multicol}
\usepackage{rotating}
\usepackage{array}

\usepackage{xurl}
\usepackage{hyperref}
\tcbuselibrary{breakable}

\usepackage[switch]{lineno}

\usepackage{graphicx}
\usepackage{dblfloatfix}

\theorembodyfont{\upshape}
\theoremheaderfont{\scshape}
\theorempostheader{:}
\theoremsep{\newline}

\title[Hindsight Bias in Clinical Temporal Reasoning]{Hindsight Bias in Clinical Temporal Reasoning: How Future Data Exposure Affects Large Language Model Judgment}

\author[Kumar et al.]{%
  \Name{Misaki Matsuura \nametag{$^{1}$}}
  \and
  \Name{Sayantan Kumar\nametag{$^{2}$}}
  \and
  \Name{Ojas Kadam\nametag{$^{3}$}}
  \and
  \Name{Jeremy C. Weiss\nametag{$^{2}$}}
  \\[3pt]
\addr{$^{1}$Case School of Engineering, Case Western Reserve University, USA} \\
\addr{$^{2}$National Library of Medicine, National Institutes of Health, USA}
  \\
  \addr{$^{3}$Rice University University, USA}
}

\begin{document}

\maketitle

\ifmlhdemo\else

\begin{abstract}
Clinical decisions are prospective, but clinical language models are often evaluated on retrospective records that reveal the final diagnosis, treatment response, and outcome. Such evaluations may reward the use of future information rather than reasoning under the uncertainty present at the decision point. We introduce a paired benchmark for measuring outcome-conditioned shifts consistent with hindsight bias in clinical temporal reasoning. It contains 171 case reports from the PubMed Central Open Access Subset---40 sepsis and 131 GLP-1/diabetes cases---represented as both textual narratives and human-annotated and LLM-generated textual time series (TTS). For each case, questions are tied to a clinically meaningful cutoff and paired with a prospective reference answer and an outcome-consistent \emph{hindsight trap}. Models answer each question using either a TTS truncated at the cutoff or the complete timeline; additional conditions vary the narrative source (original or synthetic) and TTS annotation source (human or LLM). We evaluate accuracy (Acc), hindsight trap rate (HTR), answer instability rate (AIR), and hindsight bias rate (HBR), each of which captures different signals of hindsight bias. Across GPT 5.6 Sol, Gemma 4, GLM 5.2, and Opus 5, full timeline exposure produces consistent hindsight-sensitive shifts, while temporal masking reduces bias without lowering accuracy.
\end{abstract}

\fi
\ifmlhneedsstatements
\paragraph*{Data and Code Availability}
The source case reports are publicly available through the PubMed Central Open Access Subset. Code will be provided as supplementary material.

\paragraph*{Institutional Review Board (IRB)}
This study uses publicly available case reports and involves no participant recruitment, intervention, or access to private identifiable data. 
\fi
\begin{figure*}[t]
    \centering
    \includegraphics[width=\textwidth]{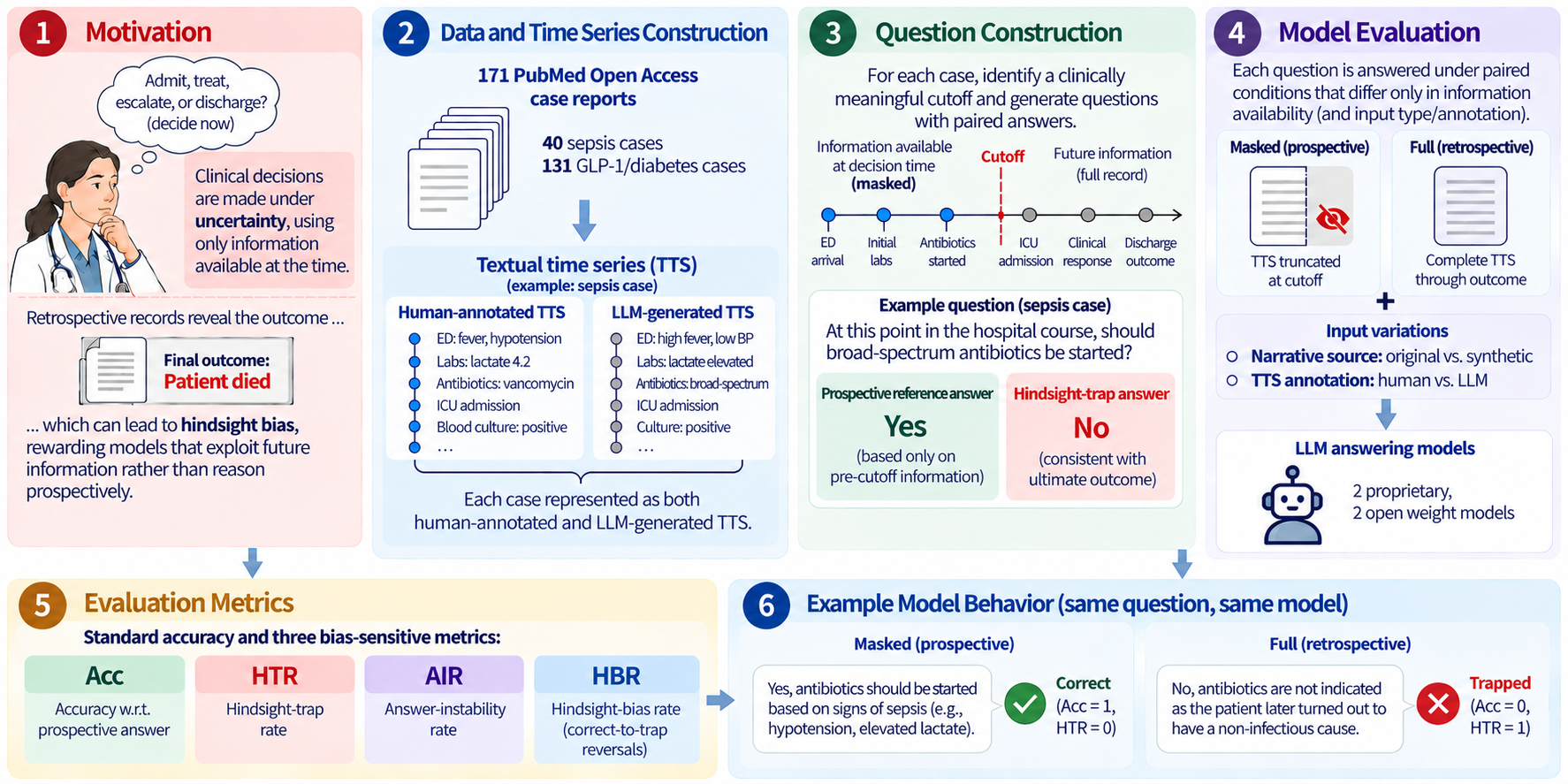}
    \vspace{-5pt}
    \caption{Benchmark pipeline for measuring hindsight-sensitive clinical reasoning. Models answer the same question using masked and complete case trajectories, and response changes are evaluated for correctness, instability, and movement toward an outcome-consistent hindsight trap.}
    \label{fig:pipeline_fig}
    \vspace{-20pt}
\end{figure*}
\section{Introduction}
\label{sec:intro}

Clinical decision-making is inherently prospective. A clinician deciding whether to admit, treat, escalate, or discharge a patient must act using only the evidence available at that moment, before the subsequent trajectory and outcome are known. Clinical artificial intelligence systems inherit the same information constraint: a model intended to support a decision at time $t$ should be evaluated using information available by time $t$, rather than information documented later in the episode.

This evaluation setting creates the conditions for hindsight bias, in which knowledge of an outcome changes judgments about what was knowable or predictable beforehand \citep{fischhoff1975hindsight}. Hindsight bias has been observed in physicians estimating diagnostic probabilities \citep{arkes1981hindsight,dawson1988hindsight} and evaluating earlier care. In a randomized vignette study, clinicians judged identical antecedent care to recovery or mortality outcomes more harshly after a death outcome in two of three cases, and greater seniority did not eliminate the effect \citep{banhamhall2019hindsight}.

However, not every answer change after additional evidence is a bias. If the task asks for the final diagnosis, later evidence should appropriately update the answer. The narrower failure mode studied here arises when later information changes a judgment whose intended target is what was supportable at an earlier reference point. Measuring this failure requires holding the clinical question and antecedent evidence fixed while varying access to post-reference information. Prior paired clinical vignette studies have shown that general and reasoning-focused LLMs can exhibit hindsight bias when outcome cues are manipulated, but these evaluations are based on individual vignettes rather than a corpus of longitudinal clinical cases \citep{wang2024cognitivebias,degany2025o1bias}. ExAnte evaluates temporal leakage in general domain QA by imposing explicit temporal cutoffs \citep{liu2026exante}, but its objective differs from ours. It evaluates whether answers are supported by information available before a specified time, whereas we ask whether exposure to a patient's subsequent clinical trajectory changes a model's judgment about what was supportable at an earlier decision point. This requires paired full--masked evaluation with a fixed prospective reference answer and an outcome-consistent hindsight trap, allowing us to measure not only correctness but the direction of within-question answer changes. To the best of our knowledge, no existing clinical benchmark applies this design across a corpus of real longitudinal case trajectories. 



\vspace{-1.5mm}
\paragraph{Contributions.}
We introduce (1) a paired benchmark for clinical hindsight bias that
compares masked and complete trajectories using prospective reference
answers and predefined hindsight traps; (2) complementary metrics
($\Delta$HTR, AIR, and HBR) for measuring directional
hindsight-sensitive changes; and (3) a cross-model evaluation across
two clinical cohorts and multiple input representations showing that
temporal masking reduces hindsight bias without reducing accuracy.

\vspace{-3mm}
\section{Related Work}
\label{sec:related_work}
\vspace{-1mm}

Here, we present an overview of related works. A detailed version can be found in Appendix~\ref{app:lit_review_detailed}.

\paragraph{General temporal reasoning.}
TRAM, TimeBench, Test of Time, and TimE evaluate event ordering, temporal arithmetic, duration, and multistep reasoning in synthetic or general domain settings \citep{wang2024tram,chu2024timebench,fatemi2025testoftime,wei2025time}. ExAnte is closest to our temporal setup. It imposes explicit cutoffs and measures leakage from post-cutoff knowledge, but its tasks are non-clinical \citep{liu2026exante}. Our benchmark instead tests whether later events in a patient trajectory shift a judgment about an earlier clinical point toward the hindsight trap.

\vspace{-1mm}
\paragraph{Clinical question answering over longitudinal records.}
MedAlign, MIMIC-Instr, EHRNoteQA, and EHRSQL evaluate instruction following or question answering over longitudinal notes and structured EHR data \citep{fleming2024medalign,wu2024mimicinstr,kweon2024ehrnoteqa,lee2022ehrsql}. More recent benchmarks target temporal reasoning directly. TIMER grounds instructions to patient-specific timestamps, ASCENT evaluates diagnosis as evidence accumulates, and RealICU evaluates windowed ICU decisions using hindsight-informed labels \citep{cui2025timer,choi2026ascent,shen2026realicu}. These tasks measure temporal grounding or stepwise reasoning. We pair the same question and pre-cutoff evidence with complete and masked timelines, 
allowing measurement of
whether the answer changes and whether it moves in an outcome-consistent direction.

\vspace{-1mm}
\paragraph{Reasoning from clinical case reports.}
MedCaseReasoning evaluates diagnostic answers and reasoning derived from open access case reports, while MedR-Bench covers examination, diagnosis, and treatment planning \citep{wu2025medcasereasoning,qiu2025medrbench}. Prior work also reconstructs time-localized sepsis trajectories from PubMed Central case reports as textual time series \citep{noroozizadeh2026sepsistts}. We build on this representation but study a different question: not only whether a model reaches the final conclusion, but whether that conclusion was supportable before later events were known.


\vspace{-3mm}
\section{Methods}
\label{sec:methods}
\vspace{-1mm}

The overall pipeline is illustrated in Figure~\ref{fig:pipeline_fig}, and all prompts used are listed in Appendix~\ref{app:prompts}.

\subsection{Study design and data source}
\label{sec:data}
\vspace{-1mm}
Case reports were drawn from the PubMed Central Open Access (PMC OA) Subset provided by \cite{nlm2025pmcoa}. We used two cohorts comprising 171 case reports: 40 sepsis cases and 131 cases of type 2 diabetes patients taking Glucagon-like Peptide Receptor Agonists (GLP1-RA) medications. The sepsis and GLP1-RA cohorts build on previously published works \citep{noroozizadeh2026sepsistts,kumar2026glp1ra}, which provide independently constructed timelines from clinically trained annotators and LLM extractors. 
The two cohorts were selected to evaluate the benchmark across contrasting clinical trajectories: comparatively dense, acute care episodes in sepsis and more heterogeneous longitudinal trajectories in GLP-1/diabetes. 
\vspace{-2mm}
\subsection{Clinical representations}
\label{sec:representations}
\vspace{-1.5mm}
Each case $i$ was represented by its original case report narrative $N_i$ and by two textual time series (TTS): a human-annotated TTS $T_i^{\mathrm{H}}$ and an independently generated LLM TTS $T_i^{\mathrm{L}}$. A TTS is a sequence $T_i^{s}=\{(e_{ij}^{s},t_{ij}^{s})\}_{j=1}^{n_i^{s}}, \qquad s\in\{\mathrm{H},\mathrm{L}\}$,
where $e_{ij}^{s}$ is a clinical event and $t_{ij}^{s}$ is its relative time in hours. This representation preserves the distinction between narrative order and clinical event order, which may diverge in retrospectively written case reports.


\vspace{-1mm}
 

\vspace{-2mm}
\subsection{Question construction}
\label{sec:question_generation}
\vspace{-1.5mm}

Questions were generated with GPT~5.6 Sol with the default medium reasoning effort from the original narrative $N_i$ and human TTS $T_i^{\mathrm{H}}$. For each case, the generator returned ten questions as a schema-constrained JSON array; outputs that failed validation received one repair attempt. Across 171 cases, this yielded
1,710 questions. Figure~\ref{fig:question_example} shows a representative item, including the fields visible to the answering model and the hidden metadata used for evaluation.

\begin{figure}[t]
\centering
\fbox{%
\begin{minipage}{0.95\columnwidth}
\footnotesize

\textbf{Case:} 58-year-old male admitted with septic shock secondary to pneumonia. [...]

\medskip
\textbf{Cutoff time:} 24 hours post-admission

\medskip
\textbf{Visible to answering model:}

\smallskip
\textit{Question:} What probability range best represents the likelihood of requiring mechanical ventilation within 48 hours, given the patient's current respiratory status and haemodynamic profile?

\smallskip
\textit{Format:} Multiple choice (ordinal)

\smallskip
\textit{Options:} [A.] $<$10\%, [B.] 10--30\%, [C.] 31--60\%, [D.] $>$60\%

\noindent\rule{\linewidth}{0.4pt}

\textbf{Hidden evaluation metadata:}

\smallskip
$a^*$ \textbf{(reference answer):} A. $<$10\%

\smallskip
$a^{\mathrm{trap}}$ \textbf{(hindsight trap):} D. $>$60\%

\smallskip
\textbf{Trap explanation:} The patient was intubated at 36 hours; a model with access to the full trajectory may inflate the prospective probability estimate in light of this outcome.

\smallskip
\textbf{Eval target:} \texttt{both}

\smallskip
\textbf{Cutoff event:} Arterial blood gas showing mild hypoxaemia; vasopressor requirement stable.

\end{minipage}%
}
\caption{Representative benchmark question. The visible component (above the divider) is shown to the answering model; the hidden metadata (below) is used only for evaluation. The reference answer $a^*$ is what was defensible at the cutoff, while the hindsight trap answer $a^{\mathrm{trap}}$ is what may seem more compelling once the eventual outcome is known.}
\label{fig:question_example}
\vspace{-8.5mm}
\end{figure}


Questions were assigned to \texttt{accuracy}, \texttt{hindsight\_bias}, or \texttt{both}. \texttt{accuracy} items targeted verifiable answers
with low hindsight sensitivity; \texttt{hindsight\_bias} items targeted subjective
judgments such as likelihood and treatment appropriateness that could shift after later evidence; and \texttt{both} items required a
prospectively supported answer and a plausible trap. Each case targeted a 2/3/5 split across these categories.

The benchmark included free response, true/false, and four-option multiple choice questions, with multiple choice items labeled categorical or ordinal. Ordinal options used explicit ranges, with $a_q^*$ and $a_q^{\mathrm{trap}}$ placed at maximally separated positions. The prompt limited each reasoning type to two questions per case to maximize diversity. It also omitted explicit temporal references from visible question text so that prospective reasoning depended on the supplied context rather than an overt instruction.

\vspace{-2mm}
\subsection{Temporal masking and cutoff position}
\label{sec:masking}
\vspace{-1.5mm}

For each question $q$, the generator supplied a numeric cutoff $c_q$ in
the time coordinate of the human TTS. Given a TTS $T_i^s$, we retained
events at or before the cutoff: 
\setlength{\abovedisplayskip}{1pt}
\setlength{\belowdisplayskip}{2pt}
\begin{equation}
T_{iq,\mathrm{masked}}^{s} = \{(e_{ij}^{s},t_{ij}^{s})\in T_i^{s}:t_{ij}^{s}\leq c_q\}
\end{equation}
while $T_{i,\mathrm{full}}^{s}=T_i^s$ retained the complete trajectory.
The same numeric cutoff was applied to the human and LLM TTS.

After sorting valid rows by numeric time and
preserving source order within ties, let
$k_q\in\{1,\ldots,n_i\}$ denote the chronological index matching the
cutoff. We define $\operatorname{RelPos}(q)=\frac{k_q-1}{n_i-1},$
where 0 and 1 denote the first and last recorded events, respectively.

\paragraph{Constructing synthetic case reports.} 
To test whether our results depended on the original case report prose or persisted under an independently reconstructed narrative representation, we generated synthetic narratives from the TTS. A separate narrative was generated for every question $q$, cutoff $c_{iq}$, and annotation source $s \in \{\mathrm{H}, \mathrm{L}\}$: each was derived from the masked TTS $T_{iq}^{s,\mathrm{mask}}$ rather than the full TTS, so no post-cutoff information could leak into the narrative seen by the answering model. A single case report $i$ with $Q_i$ questions and two annotation sources therefore yields up to $2Q_i$ distinct synthetic narratives, guaranteeing strict temporal alignment between the narrative and the TTS in each paired condition. For each condition, narratives were generated with the same model used for answering.




\vspace{-2mm}
\subsection{Answer generation}
\label{sec:answering}
\vspace{-1.5mm}

We evaluated GPT-5.6 Sol, Gemma-4-31B, GLM-5.2-FP8, and Opus~5. Each model
answered every question independently and received only the visible
question fields---\texttt{question}, \texttt{format}, and
\texttt{options}---together~ with the context specified by
Table~\ref{tab:conditions}. The rest were
withheld; see Figure~\ref{fig:question_example}. Responses contained an \texttt{answer} and brief supporting
\texttt{evidence}. No model had access to external retrieval, tools, or
the web. 

\paragraph{Hindsight warning sensitivity analysis.}
We repeated the Sepsis experiments for GPT and GLM after appending
explicit instructions to use only contemporaneously available
information and disregard the final outcome. All other procedures were
unchanged.

\vspace{-2mm}
\subsection{Experimental conditions}
\label{sec:conditions}
\vspace{-1.5mm}

We evaluated seven input conditions (Table~\ref{tab:conditions}, Appendix \ref{app:experimental_conditions}) using
the same question set throughout. Condition~0 provided the complete original narrative as an unpaired retrospective baseline. Conditions~1.x
combined that narrative with full or masked human- or LLM-generated TTS; only the TTS differed within each pair, so the narrative could still reveal post-cutoff information. Conditions~2.x used TTS alone and therefore directly isolated the effect of temporal masking.
Conditions~3.x paired each TTS with a synthetic narrative generated from the corresponding masked or full information scope. 

Condition~0 is unpaired and therefore supports
only Acc and HTR. Conditions~1.x, 2.x, and 3.x provide matched masked
and full responses and additionally support $\Delta$Acc, $\Delta$HTR,
AIR, and HBR.

\begin{table}[t]
\centering
\scriptsize
\setlength{\tabcolsep}{3pt}
\renewcommand{\arraystretch}{0.85}
\caption{Input conditions. H/L-TTS denote human/LLM-generated textual
time series; F/M denotes full/masked variants.}
\vspace{-2mm}
\label{tab:conditions}
\resizebox{0.95\columnwidth}{!}{%
\begin{tabular}{@{}cl@{}}
\toprule
\textbf{Condition} & \textbf{Model input} \\
\midrule
0   & Original narrative \\
1.1 & Original narrative + H-TTS (F/M) \\
1.2 & Original narrative + L-TTS (F/M) \\
2.1 & H-TTS only (F/M) \\
2.2 & L-TTS only (F/M) \\
3.1 & Synthetic narrative + H-TTS (F/M) \\
3.2 & Synthetic narrative + L-TTS (F/M) \\
\bottomrule
\end{tabular}%
}
\vspace{-3mm}
\end{table}

\vspace{-2mm}
\subsection{Evaluation}
\label{sec:evaluation}
\vspace{-1.5mm}

Let $a_{q,z}$ denote the model answer to question $q$ under visibility
condition $z\in\{\mathrm{full},\mathrm{masked}\}$. We compare each
answer with the prospective reference $a_q^*$ and the hindsight trap
$a_q^{\mathrm{trap}}$. Metrics were computed per question and then averaged. Structured responses were scored by exact match. Free responses were scored by a GPT-5.6 Sol judge
against the prospective reference and hindsight-trap answers using
$\{0,0.5,1\}$ semantic-agreement scores. 

\vspace{-2mm}
\subsubsection{Individual metrics}
\vspace{-1.5mm}

\noindent \textbf{Accuracy (Acc).} Acc measures agreement with the prospective reference answer $a_q^*$
and is defined for \texttt{accuracy} and \texttt{both} questions,
which comprise 70\% of the dataset. For true/false and multiple choice
questions, $\mathrm{Acc}_q(a) = \mathbf{1}[a = a_q^*] \in \{0, 1\}$,
whereas free response questions receive
$\mathrm{Acc}_q(a) \in \{0,0.5,1\}$ from the judge, with 0.5 denoting
partial correctness.

\noindent \textbf{Hindsight trap rate (HTR).} HTR measures agreement with the outcome-consistent hindsight trap
$a_q^{\mathrm{trap}}$ and is defined for
\texttt{hindsight\_bias} and \texttt{both} questions, which comprise
80\% of the dataset. For true/false and categorical multiple choice
questions,
\setlength{\abovedisplayskip}{1pt}
\setlength{\belowdisplayskip}{2pt}
\begin{equation}
\mathrm{HTR}_q(a) = \mathbf{1}[a = a_q^{\mathrm{trap}}] \in \{0, 1\},
\end{equation}
whereas free response questions receive
$\mathrm{HTR}_q(a) \in \{0,0.5,1\}$ from the judge. For ordinal
multiple choice questions, HTR is graded by movement from $a_q^*$
toward $a_q^{\mathrm{trap}}$. In
Figure~\ref{fig:question_example}, for example, answers B, C, and D
receive $\tfrac{1}{3}$, $\tfrac{2}{3}$, and 1, respectively, yielding
$\mathrm{HTR}_q(a)\in\{0,\tfrac{1}{3},\tfrac{2}{3},1\}$.

\vspace{-2mm}
\subsubsection{Paired metrics}
\vspace{-1.5mm}

\noindent \textbf{Delta accuracy ($\Delta$Acc).}
\setlength{\abovedisplayskip}{2pt}
\setlength{\belowdisplayskip}{2pt}
\begin{equation}
\Delta\mathrm{Acc}_q =
\mathrm{Acc}_q(a_{q,\mathrm{masked}})
-
\mathrm{Acc}_q(a_{q,\mathrm{full}}).
\end{equation}
A positive $\Delta\mathrm{Acc}_q$ indicates greater correctness under
masking, meaning
post-cutoff information reduces average accuracy.

\noindent \textbf{Delta hindsight trap rate ($\Delta$HTR).}
\setlength{\abovedisplayskip}{2pt}
\setlength{\belowdisplayskip}{2pt}
\begin{equation}
\Delta\mathrm{HTR}_q =
\mathrm{HTR}_q(a_{q,\mathrm{full}})
-
\mathrm{HTR}_q(a_{q,\mathrm{masked}}).
\end{equation}
A positive $\Delta\mathrm{HTR}_q$ indicates greater trap attraction
under full exposure, meaning post-cutoff
information increases hindsight bias.

\noindent \textbf{Answer instability rate (AIR).}
AIR records any answer change, regardless of direction:
\setlength{\abovedisplayskip}{2pt}
\setlength{\belowdisplayskip}{2pt}
\begin{equation}
\mathrm{AIR}_q =
\mathbf{1}
[a_{q,\mathrm{full}} \not\equiv a_{q,\mathrm{masked}}]
\in \{0, 1\}.
\end{equation}
AIR upper-bounds any
bias effect because a model answer must change to exhibit
hindsight bias.

\noindent \textbf{Hindsight bias rate (HBR).}
HBR captures the joint event that a model is fully correct under
masking and moves toward the trap under full exposure. It is defined
only for \texttt{both} questions, for which both $a_q^*$ and
$a_q^{\mathrm{trap}}$ are available. Let
$I_q^{\mathrm{M}} =
\mathbf{1}[\mathrm{Acc}_q(a_{q,\mathrm{masked}}) = 1]$. Then
\setlength{\abovedisplayskip}{2pt}
\setlength{\belowdisplayskip}{2pt}
\begin{equation}
\mathrm{HBR}_q =
I_q^{\mathrm{M}}
\cdot
\mathrm{HTR}_q(a_{q,\mathrm{full}}),
\label{eq:hbr}
\end{equation}
so HBR is nonzero only when the masked answer is correct.

\subsubsection{Statistical analyses}
We estimated 95\% confidence intervals using a case-clustered
bootstrap with 2{,}000 replicates. Cases were sampled with replacement,
and all questions and paired responses from each sampled case were
retained. Sepsis and
GLP-1/diabetes were analyzed separately. For metrics evaluated relative to zero, effects were considered statistically significant when the 95\% bootstrap confidence interval excluded zero.

\subsection{Human validation}
\label{sec:human_validation}


Four clinical annotators (one physician and three medical or physician
assistant trainees) reviewed an error-enriched 10\% sample of cases,
split into two non-overlapping subsets with two annotators each.
Annotators assessed whether each question was clinically reasonable
and whether they agreed with the prospective reference and hindsight-trap
answers. Sampling and validation details are provided in Appendix~\ref{app:validation}.




\vspace{-3mm}
\section{Results}
\label{sec:results}
\vspace{-1mm}

\begin{figure*}[t]
    \centering
    \includegraphics[width=0.999\textwidth]{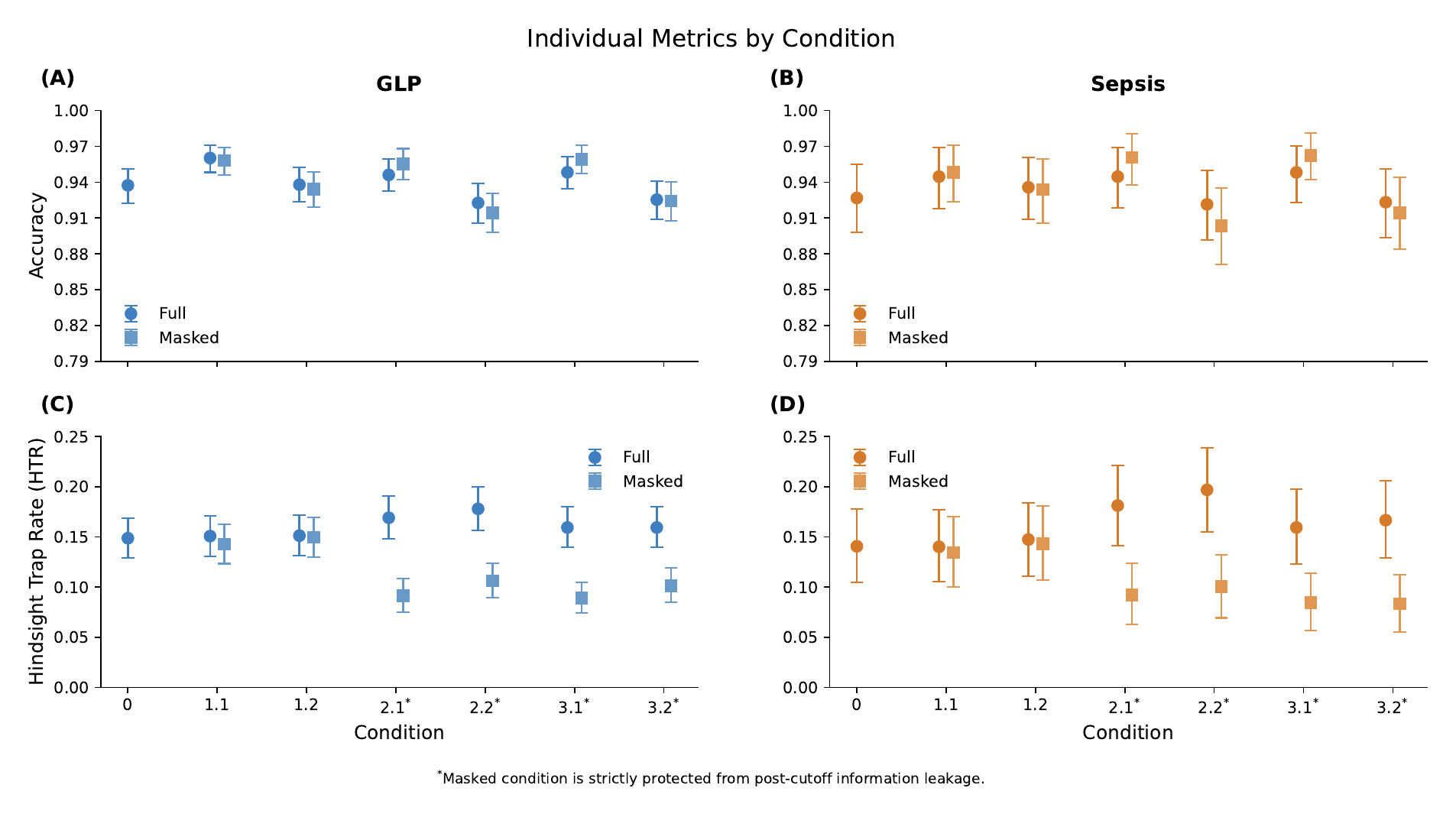}
    \vspace{-2mm}
    \caption{
        Accuracy and Hindsight Trap Rate (HTR) across experimental conditions for GPT for the GLP and sepsis datasets.
        Condition 0 represents the full information setting.
        For Conditions 1.1--3.2, results are shown for both the full and masked settings.
        Points indicate observed values and error bars indicate 95\% bootstrap confidence intervals.
    }
    \label{fig:accuracy_htr}
    \vspace{-7.5mm}
\end{figure*}

Figure~\ref{fig:accuracy_htr} summarizes Accuracy and Hindsight Trap Rate (HTR) for GPT-5.6 Sol on the GLP and Sepsis datasets. Condition 0 serves as the full information baseline, while Conditions 1.1--3.2 allow direct comparison between the full and masked settings.
Results for Gemma 4, GLM 5.2, and Opus 5 are provided in Appendix~\ref{app:all_results}.


\begin{figure*}[t]
    \centering
    \includegraphics[width=0.999\textwidth]{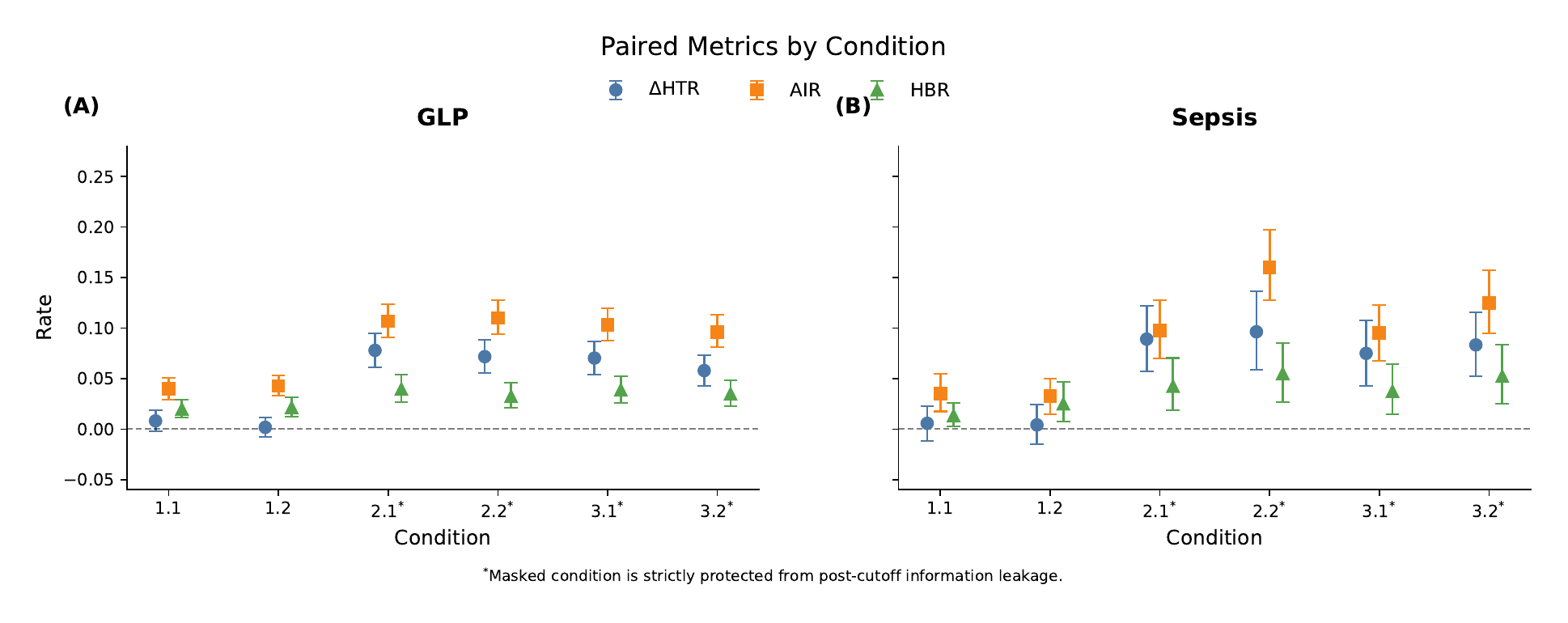}
    \vspace{-1mm}
    \caption{
        Hindsight bias metrics across paired experimental conditions for GPT for the GLP and sepsis datasets.
        We report $\Delta$HTR, AIR, and HBR for Conditions 1.1--3.2.
        Points indicate observed values and error bars indicate 95\% bootstrap confidence intervals.
    }
    \label{fig:hindsight_bias_metrics}
    \vspace{-7.5mm}
\end{figure*}

\paragraph{Naive narrative answering already shows substantial hindsight bias.}
Before considering the paired masking experiments, condition~0 establishes that hindsight bias is already a meaningful problem when the model answers questions directly from the clinical narrative. As shown in Figure~\ref{fig:accuracy_htr}C and Figure~\ref{fig:accuracy_htr}D, in both datasets, the hindsight trap rate (HTR) is significantly greater than zero. 14--15\% of responses are hindsight-biased when naively answering questions based on narratives.

\paragraph{Paired masking reveals robust hindsight bias effects.}
The paired conditions (1.x--3.x) allow hindsight bias to be quantified
directly using three complementary metrics: $\Delta$HTR, AIR, and HBR.
As seen in Figure~\ref{fig:hindsight_bias_metrics}, across both datasets,
AIR and HBR are significantly greater than zero in every paired condition,
indicating that access to future information systematically changes model
responses and, more stringently, can shift initially correct prospective
answers toward predefined hindsight traps. This is particularly notable
for HBR, which is a strict subset of AIR and therefore distinguishes
hindsight-biased shifts from general answer instability. The evidence is
strongest under strict temporal masking (Conditions~2.x and~3.x), where
all three metrics are significantly positive across both datasets. For
example, AIR reaches 16.0\% in Sepsis Condition~2.2, while HBR reaches
5.5\% in the same condition, demonstrating that a meaningful subset of
responses not only changed after future information was revealed but
shifted specifically from a prospectively correct answer toward the
hindsight trap. Overall, the consistency of $\Delta$HTR, AIR, and HBR
under strict masking shows that the observed effects are not merely
general response instability: access to future clinical information
systematically shifts model judgments toward hindsight-consistent answers.


\paragraph{Condition~1 provides weaker temporal isolation.}
In conditions~1.1 and~1.2, $\Delta$HTR remains small and non-significant despite significant AIR and HBR (Figure~\ref{fig:hindsight_bias_metrics}). Unlike conditions~2.x and~3.x, the masked variant in condition~1 retains the complete original narrative, which may contain post-cutoff or outcome-relevant information. Consequently, the full--masked comparison isolates only the effect of masking the TTS, not the effect of removing future information from the model's entire input. The significant AIR and HBR nevertheless show that
changing TTS access can alter responses, including shifts from prospectively correct answers toward hindsight traps, even when the retrospective narrative remains available.


\paragraph{Human versus LLM TTS.}
Across both datasets, the human-TTS variants generally achieve slightly higher absolute accuracy than the corresponding LLM-TTS variants. This comparison should be interpreted in light of the benchmark construction: the human TTS was used as the reference representation during question generation. Discrepancies in the LLM-reconstructed TTS---including differences in event content or timing---may therefore omit information needed to answer a question or place information on the wrong side of its masking cutoff.

\paragraph{The synthetic narrative conditions support the masking interpretation.}
Conditions~3.x are especially informative because they retain a narrative style prompt while restoring temporal control through a synthetic narrative. Their results closely track the strict masking pattern seen in 2.x: all hindsight bias metrics are significantly positive, and their magnitudes are much closer to 2.x than to the attenuated 1.x conditions. This suggests that the weaker effects in 1.x are not due to narrative prompting per se, but to leakage from the original unmasked narrative. The synthetic narrative therefore appears to preserve the benefits of a narrative format without undermining the masking manipulation.

\paragraph{Bias reduction does not come at the cost of accuracy.}
In contrast to the consistent hindsight bias signal, Figure~\ref{fig:accuracy_htr}A and Figure~\ref{fig:accuracy_htr}B highlight that $\Delta$Acc is not significant in any paired condition in either dataset. Although some point estimates are positive and others negative, there is no reliable evidence that access to the full case improves accuracy relative to the masked version. This is a central result that masking reduces hindsight bias without imposing a detectable accuracy penalty. Put differently, future information is systematically biasing the model's answers, but not meaningfully helping it answer correctly. This strengthens the case for temporally masked evaluation as a way to reduce hindsight bias while preserving task performance.

\paragraph{Explicit hindsight warning attenuates but does not eliminate the effect.}
As a sensitivity analysis, we repeated the Sepsis evaluation for GPT and GLM with explicit instructions to avoid hindsight. The warning generally reduced hindsight-related metrics, particularly under strict masking, while accuracy remained broadly similar. However, AIR and HBR remained positive across paired conditions, indicating that hindsight-sensitive response changes persisted despite the explicit instruction. Full results are reported in Appendix~\ref{app:modified_prompt}.

\paragraph{Human validation supports benchmark label validity.}
Human review of the error-enriched validation sample found that
95.9\% of questions were judged clinically reasonable, with agreement
rates of 98.7\% for the prospective reference answers and 97.1\% for the
hindsight-trap answers. Inter-annotator agreement was 91.8\% for question
reasonableness, 99.2\% for the prospective answer, and 94.1\% for the hindsight
trap. Detailed results are
reported in Appendix~\ref{app:validation}.

\vspace{-3mm}
\section{Discussion}
\label{sec:discussion}
\vspace{-1mm}

\paragraph{Principal findings.}
Across both cohorts, exposure to the complete timeline shifted model responses toward outcome-consistent alternatives. In the unpaired narrative only baseline, the hindsight trap rate was 14--15\%; because this condition has no masked counterpart, it is best interpreted as a descriptive baseline rate rather than a causal estimate of hindsight bias. The paired conditions provide stronger evidence. AIR and HBR were positive across all paired input settings, and strict temporal masking (conditions~2.x and~3.x) produced the largest
changes in trap attraction, with $\Delta$HTR ranging from 5.8\% to 9.6\% and HBR from 3.3\% to 5.5\%. By contrast, $\Delta$Acc was statistically insignificant. Thus, later information changed a subset of answers in the outcome-consistent direction without a corresponding improvement in overall accuracy.

The condition comparisons clarify the role of information control. In conditions~1.x, the original narrative remained complete in both
variants, so future information was shared across the pair and the masked--full contrast was attenuated. Conditions~2.x removed the
narrative, while conditions~3.x paired each TTS with a synthetic narrative generated from the corresponding information scope. The
larger effects in these strictly controlled settings suggest that the information boundary, rather than narrative format alone, drives the measured difference. Human- and LLM-generated TTS produced the same qualitative pattern, although effect sizes varied by cohort and condition. While our experiments do not identify the source of the observed bias, its persistence after explicit hindsight warning prompting suggests
that instructions alone may be insufficient to eliminate it. This result suggests that input level temporal control may be more reliable than instruction alone when evaluating judgments about an earlier clinical state.

\vspace{-1mm}
\paragraph{Validity of benchmark labels.}
Human review provides additional support for the validity of the
GPT-generated questions, prospective reference answers, and hindsight
traps (Section~\ref{sec:human_validation};
Appendix~\ref{app:validation}). Importantly, the validation set
was intentionally enriched for cases in which models made errors or
exhibited hindsight-sensitive behavior---the same cases that contribute
most directly to our evidence of hindsight bias. Positive human
validation on this challenging subset therefore strengthens confidence
that the benchmark's central findings are not driven by invalid
questions or labels.

Furthermore, our various metrics provide robustness to
annotation error by examining
hindsight bias from different angles that rely on different annotation
components: AIR requires only an answer change, Acc depends on the
prospective reference answer, HTR depends on the hindsight-trap answer,
and HBR combines both. Convergence across these complementary measures,
particularly under strict masking, makes it less likely that the
observed hindsight-bias signal is an artifact of any single annotation
label or metric definition.

\vspace{-1mm}
\paragraph{Limitations.}
Our studies have several limitations.
\textbf{First}, the reference time was hidden and questions were intentionally unanchored, so some full--masked differences may reflect changes in task interpretation rather than hindsight bias. Explicitly specifying the reference time and allowable evidence could reduce this ambiguity, but such tightly structured questions may not reflect real clinical deployments, where models may need to infer the relevant temporal context from available information.
\textbf{Second}, free-response scoring relies on an LLM judge and may
be sensitive to paraphrase or judge-specific variation. However, the
judge only compares responses with predefined reference and
hindsight-trap answers, and most questions use structured formats,
reducing reliance on LLM-based scoring.
\textbf{Third}, published case reports overrepresent unusual and educational cases, limiting generalizability to routine clinical practice. We use these datasets because they are publicly available and, importantly, provide manually annotated TTS, enabling controlled evaluation of temporal reasoning with high-quality representations of the clinical trajectories.
\textbf{Fourth}, temporal masking relies on recorded timestamps, which
may be imprecise or misplace information relative to a cutoff.
Moreover, observation time may not reflect the true onset or duration
of a clinical state.

\vspace{-1mm}
\paragraph{Future directions.}
Evaluations of earlier clinical decisions should restrict model inputs
to information available at the decision point and report directional
measures of outcome-sensitive change alongside accuracy. Future work
should extend the benchmark to routine longitudinal EHRs, larger
expert-reviewed samples, and counterfactual outcome controls, while
investigating training strategies and model architectures that better
enforce prospective information boundaries.

\vspace{-3mm}
\section{Conclusion}
\label{sec:conclusion}
\vspace{-1mm}
We introduced a paired benchmark for testing whether later clinical information changes LLM judgments about an earlier decision point.
Across two clinical cohorts and four models, complete trajectories increased movement toward outcome-consistent hindsight traps, whereas temporal masking reduced these shifts without a detectable loss in accuracy. These findings show that retrospective clinical evaluation can overstate prospective reasoning performance and motivate evaluation protocols that restrict model inputs to information available at the target decision time.



\section*{Acknowledgements}
This research was supported by the Intramural Research Program of the National Institutes of Health (NIH) and utilized the computational resources of the \href{http://hpc.nih.gov}{NIH HPC Biowulf cluster}. The contributions of the NIH author(s) are considered Works of the United States Government. The findings and conclusions presented in this paper are those of the author(s) and do not necessarily reflect the views of the NIH or the U.S. Department of Health and Human Services. 

\bibliography{references}

\clearpage
\numberwithin{equation}{section}
\numberwithin{figure}{section}
\numberwithin{table}{section}
\numberwithin{algorithm}{section}
\appendix

\begin{center}
    {\LARGE \bfseries Supplementary Material}
\end{center}
This supplementary material provides detailed information organized as follows:
\begin{itemize}
    \item \textbf{Appendix~\ref{app:lit_review_detailed}:} Detailed literature review.
    \item \textbf{Appendix~\ref{app:experimental_conditions}:} Experimental conditions and input configurations.
    \item \textbf{Appendix~\ref{app:prompts}:} Prompts used for question construction, answer generation, evaluation, and synthetic narrative generation.
    \item \textbf{Appendix~\ref{app:masking_cutoffs}:} Distribution of temporal masking cutoffs across the clinical trajectories.
    \item \textbf{Appendix~\ref{app:all_results}:} Complete results across other models (Gemma, GLM, Opus).
    \item \textbf{Appendix~\ref{app:modified_prompt}:} Sensitivity analysis with explicit hindsight warning.
    \item \textbf{Appendix~\ref{app:validation}:} Human validation of benchmark questions, prospective answers, and hindsight traps.
\end{itemize}

\section{Detailed Literature Review}
\label{app:lit_review_detailed}
\input{appendix/lit_review_detailed}

\section{Experimental conditions}
\label{app:experimental_conditions}
\input{appendix/experimental_conditions}

\newpage
\onecolumn
\section{Prompts}
\label{app:prompts}
\input{appendix/all_prompts}

\section{Distribution of Masking Cutoffs}
\label{app:masking_cutoffs}
\input{appendix/masking_cutoff}

\section{Results for Other Models}
\label{app:all_results}

\begin{figure*}[htbp]
    \centering
    \includegraphics[width=0.9\textwidth]
        {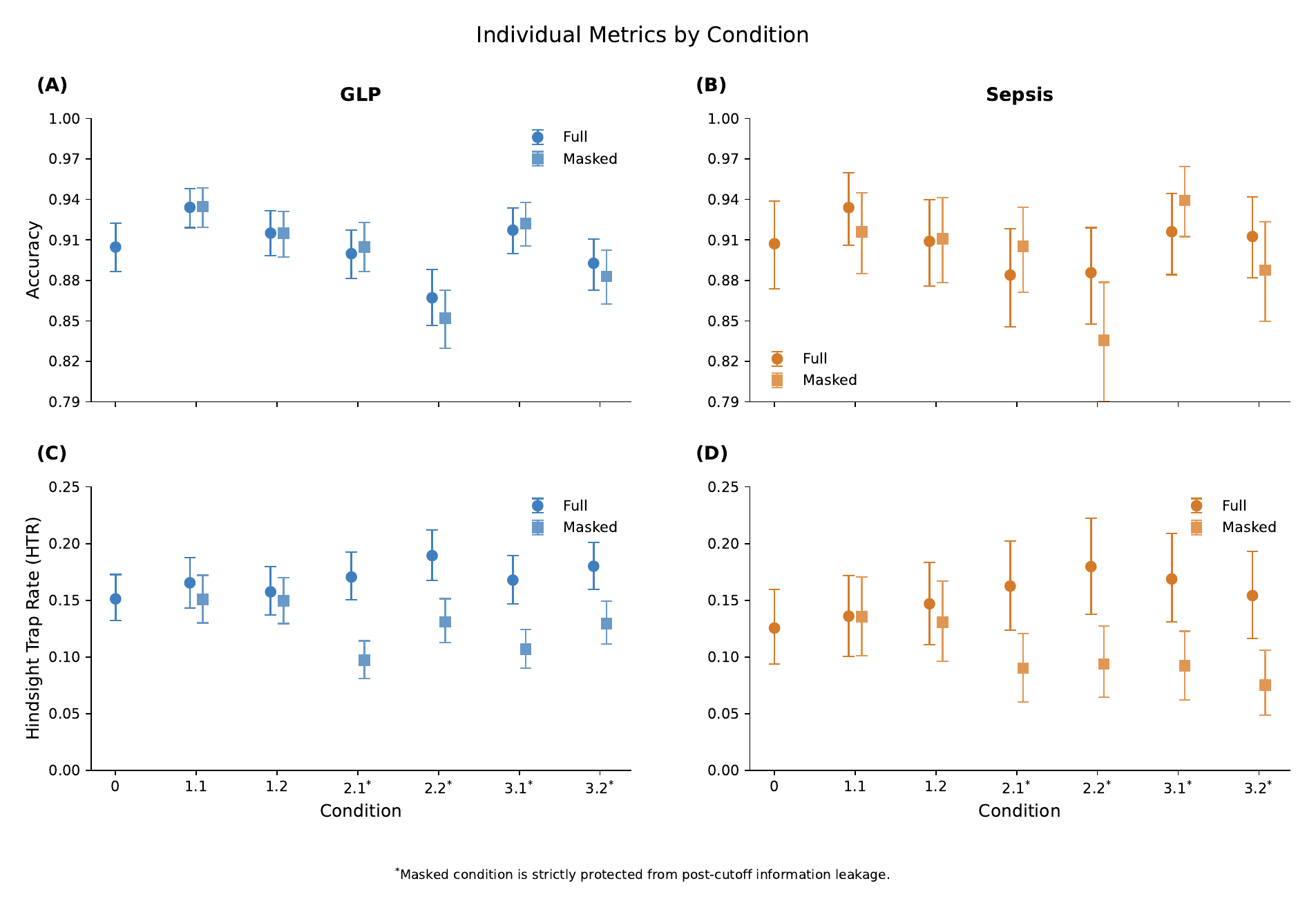}
    \caption{Accuracy and HTR results for Gemma with case-level 95\%
    bootstrap confidence intervals.}
    \label{fig:gemma_individual}
\end{figure*}

\begin{figure*}[htbp]
    \centering
    \includegraphics[width=0.9\textwidth]
        {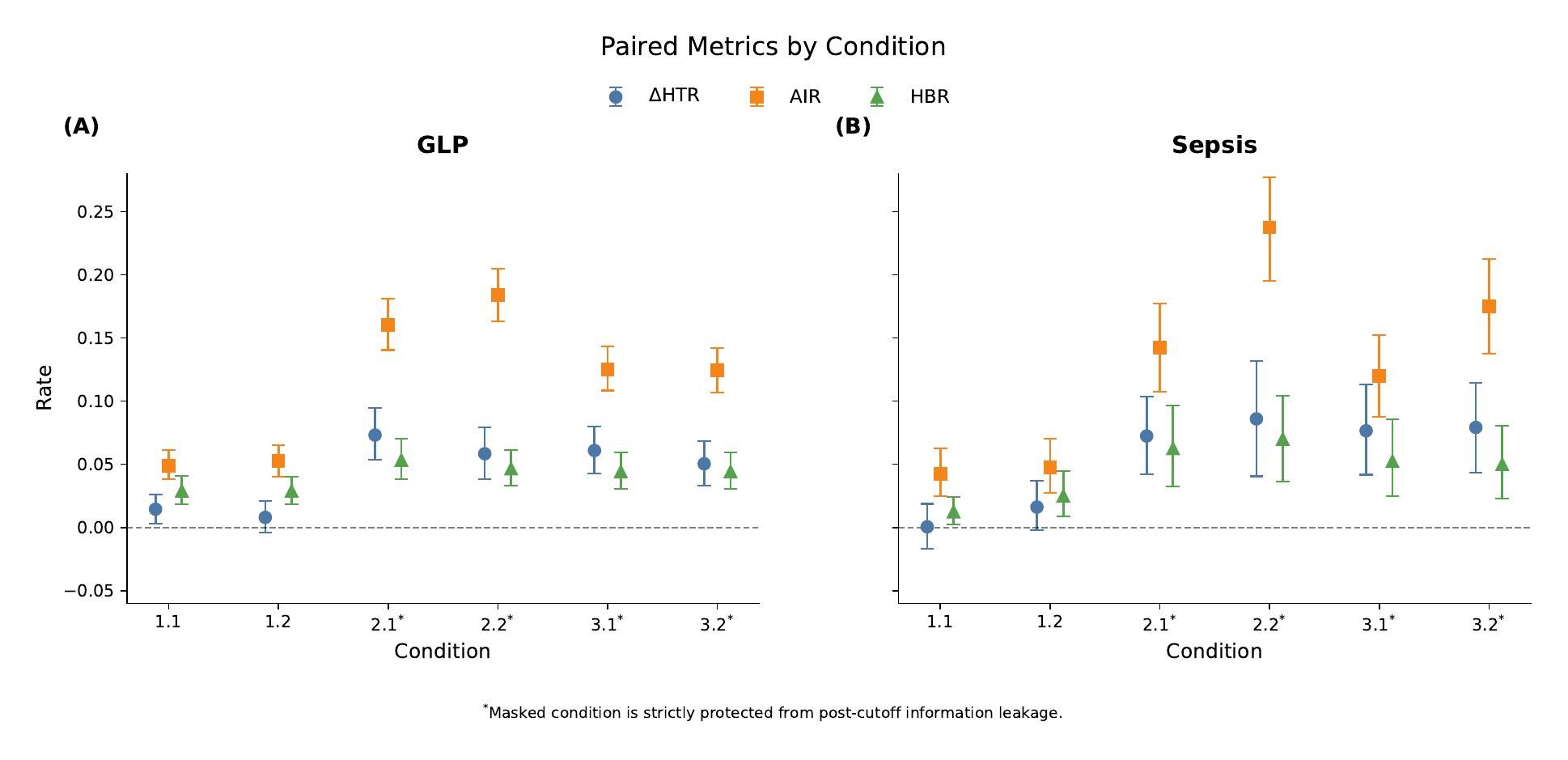}
    \caption{Hindsight bias metrics for Gemma with case-level 95\%
    bootstrap confidence intervals.}
    \label{fig:gemma_paired}
\end{figure*}

\begin{figure*}[htbp]
    \centering
    \includegraphics[width=0.9\textwidth]
        {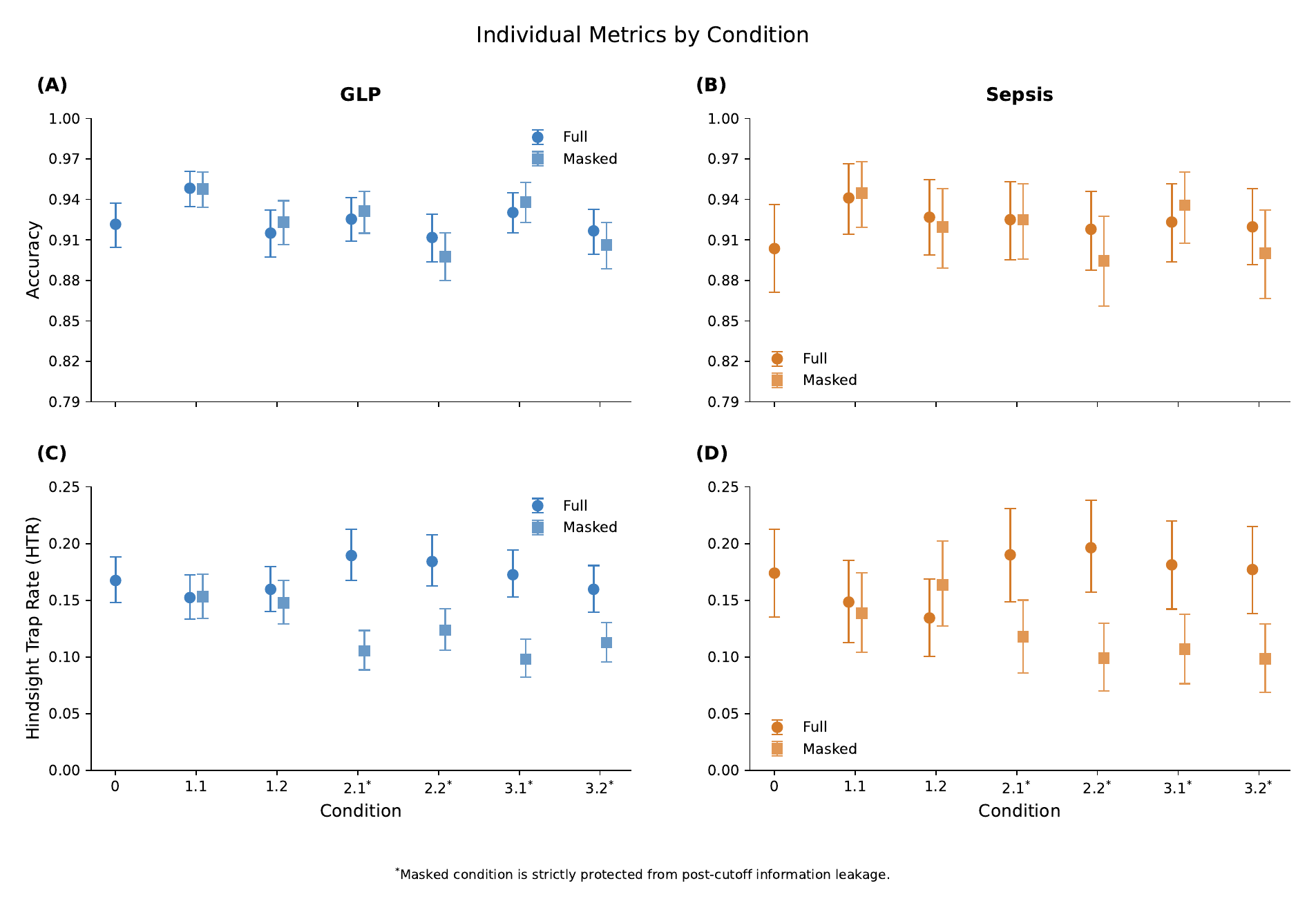}
    \caption{Accuracy and HTR results for GLM with case-level 95\%
    bootstrap confidence intervals.}
    \label{fig:glm_individual}
\end{figure*}

\begin{figure*}[htbp]
    \centering
    \includegraphics[width=0.9\textwidth]
        {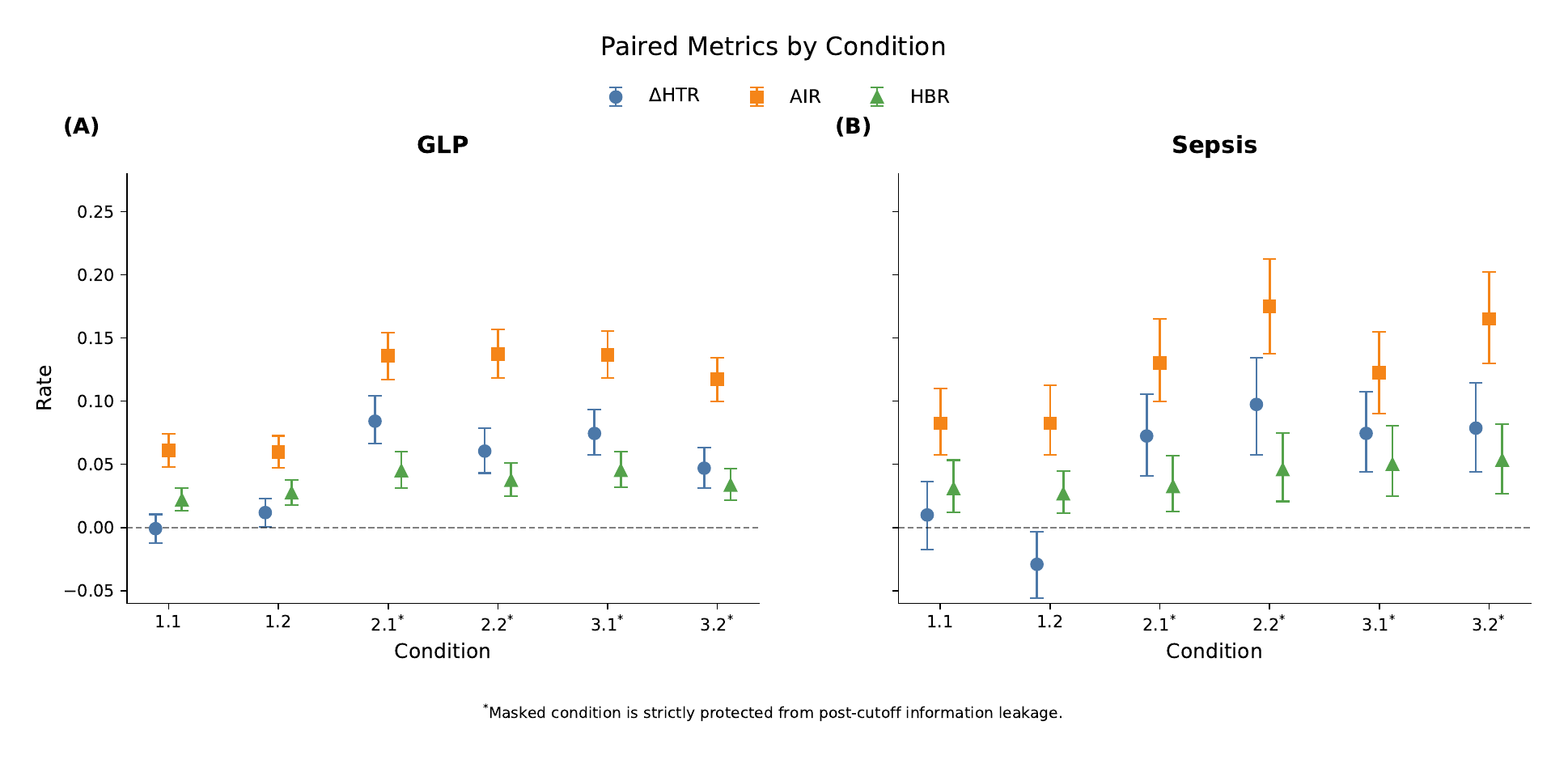}
    \caption{Hindsight bias metrics for GLM with case-level 95\%
    bootstrap confidence intervals.}
    \label{fig:glm_paired}
\end{figure*}

\begin{figure*}[htbp]
    \centering
    \includegraphics[width=0.9\textwidth]
        {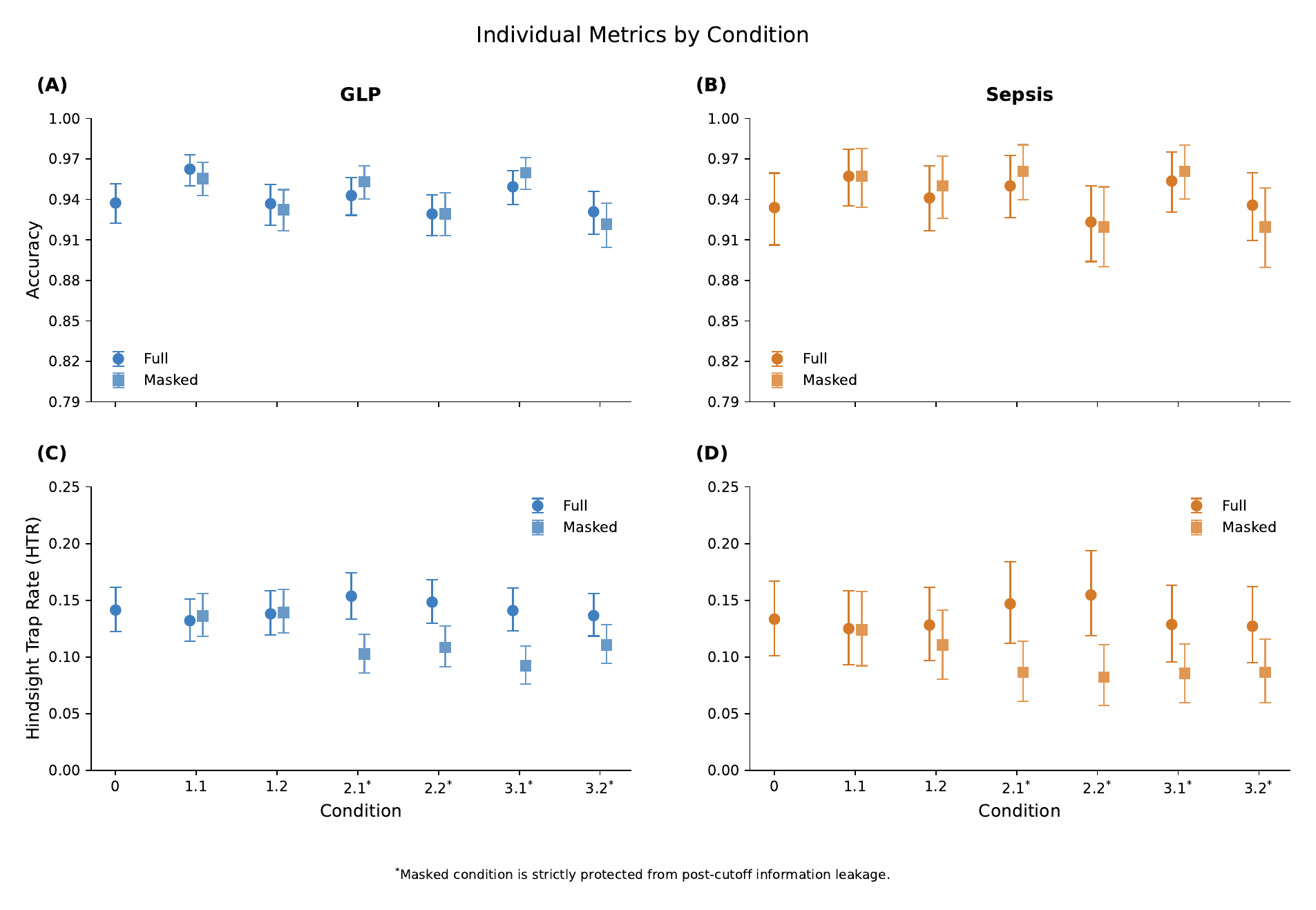}
    \caption{Accuracy and HTR results for Opus-5 with case-level 95\%
    bootstrap confidence intervals.}
    \label{fig:opus_individual}
\end{figure*}

\begin{figure*}[htbp]
    \centering
    \includegraphics[width=0.9\textwidth]
        {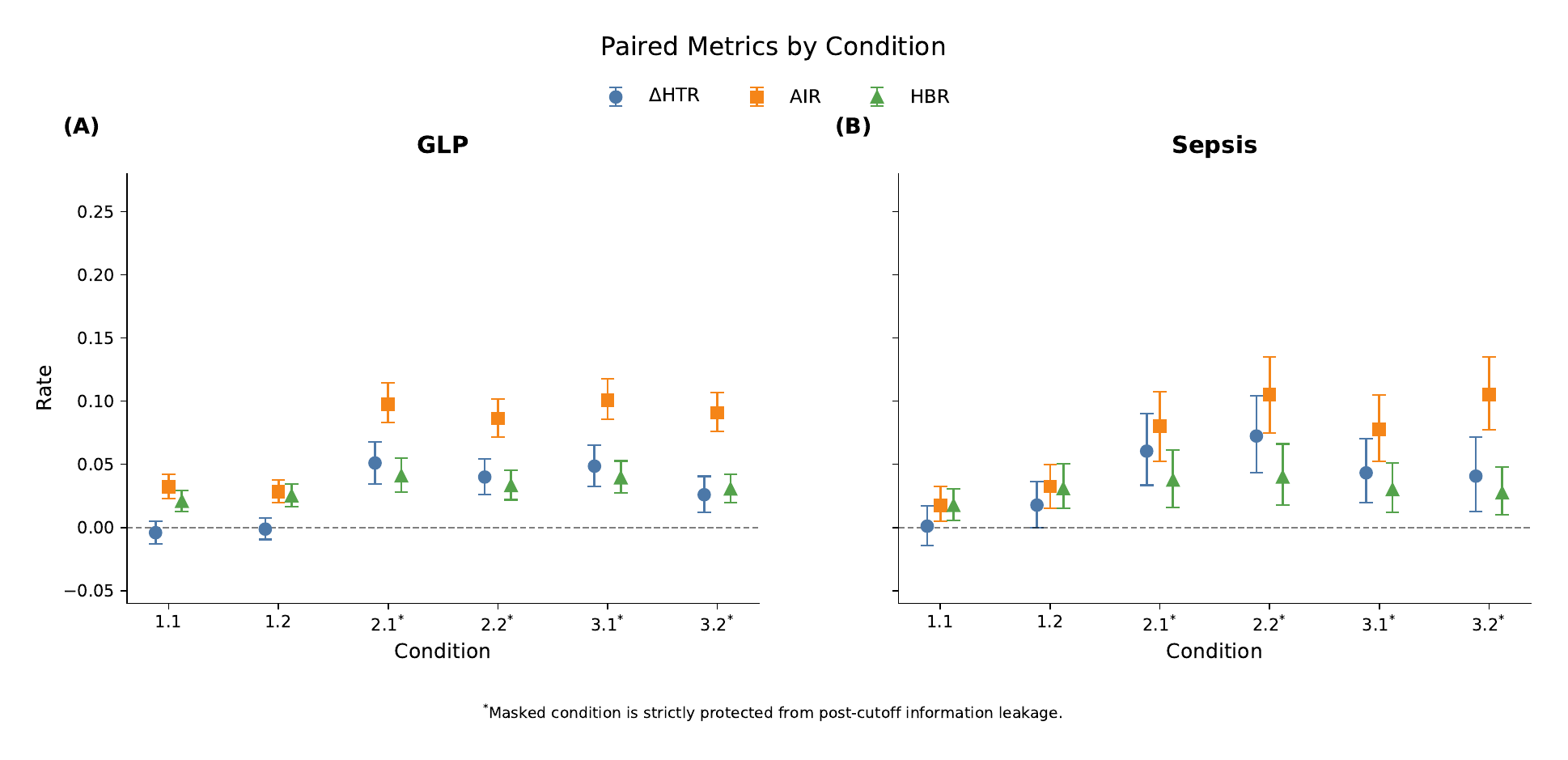}
    \caption{Hindsight bias metrics for Opus-5 with case-level 95\%
    bootstrap confidence intervals.}
    \label{fig:opus_paired}
\end{figure*}

Figures~\ref{fig:gemma_individual}--\ref{fig:opus_paired} report the
case-level 95\% bootstrap confidence intervals for the individual and
paired evaluation metrics for Gemma-4, GLM-5.2, and Opus-5. 

\begin{figure*}[t]
    \centering
    \includegraphics[width=0.8\textwidth]
    {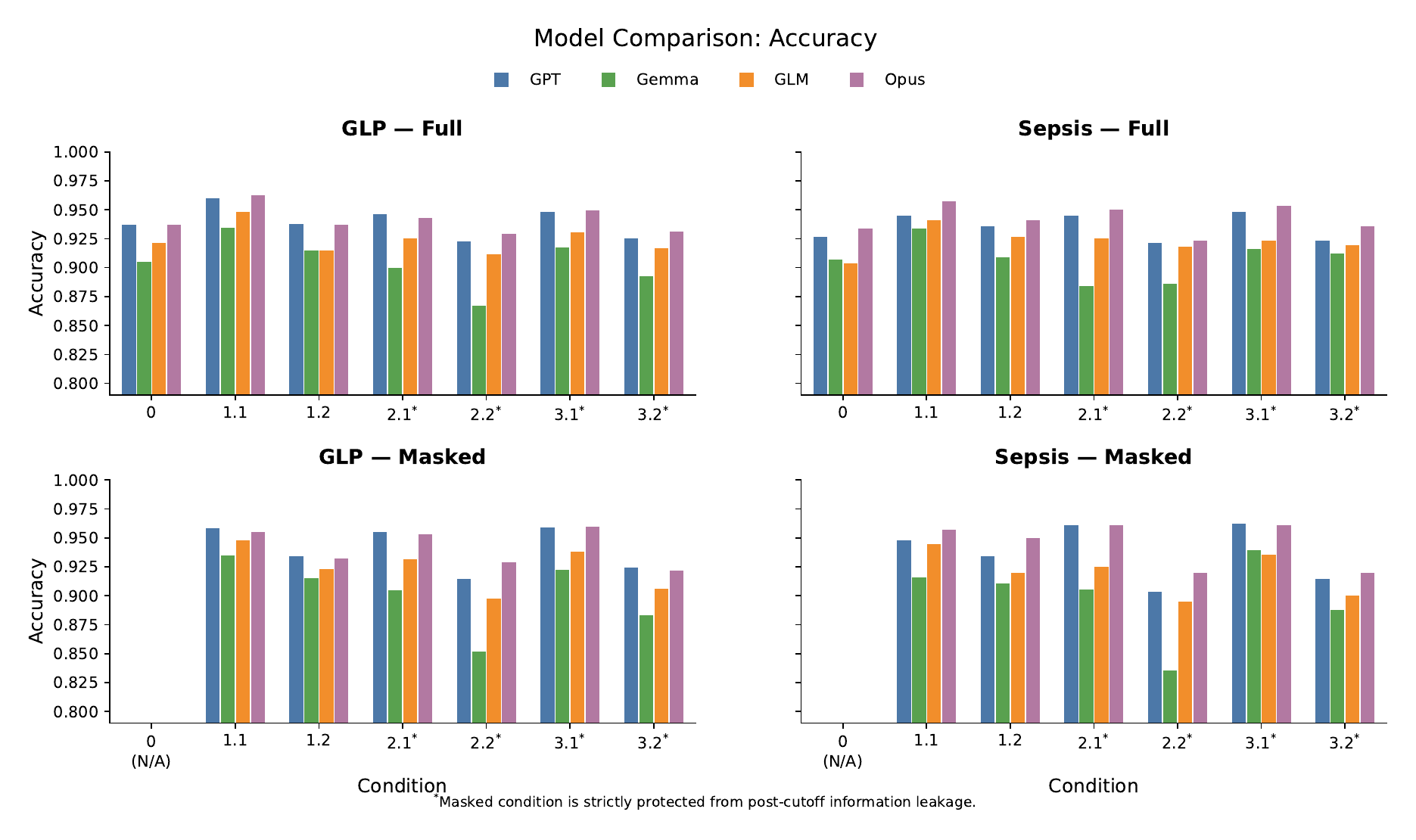}
    \caption{\textbf{Accuracy across models and input conditions.}
    Accuracy for GPT-5.6 Sol, Gemma-4, GLM-5.2, and Opus-5 on GLP and Sepsis under full
    (top) and masked (bottom) information. Asterisks denote strict
    masking conditions.}
    \label{fig:model_comparison_accuracy}
\end{figure*}

\begin{figure*}[t]
    \centering
    \includegraphics[width=0.8\textwidth]
    {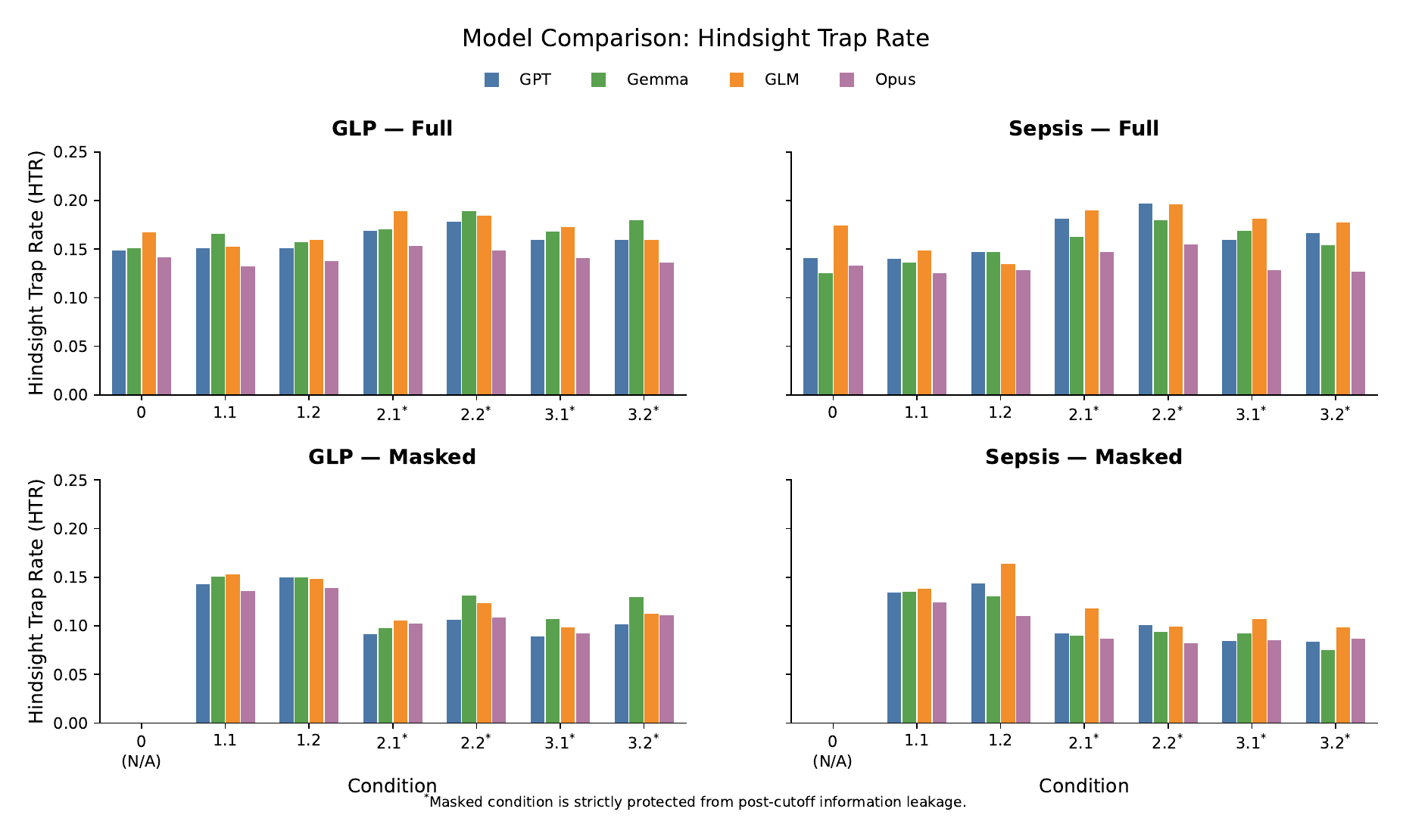}
    \caption{\textbf{Hindsight Trap Rate across models and input conditions.}
    HTR for GPT-5.6 Sol, Gemma-4, GLM-5.2, and Opus-5 on GLP and Sepsis under full
    (top) and masked (bottom) information. Asterisks denote strict
    masking conditions.}
    \label{fig:model_comparison_htr}
\end{figure*}

\paragraph{Comparison across models.}
Across GPT-5.6 Sol, Gemma-4, GLM-5.2, and Opus-5, accuracy remained generally high, with relatively modest full--masked differences for most paired conditions (Figure~\ref{fig:model_comparison_accuracy}). Condition 0 is an unpaired original-narrative baseline and is therefore shown only under full information. In contrast to accuracy, HTR showed a clearer masking effect: full and masked HTR were similar in the 1.x conditions, whereas masked HTR was consistently lower across models in the strict 2.x and 3.x conditions (Figure~\ref{fig:model_comparison_htr}).

\begin{figure*}[t]
    \centering
    \includegraphics[width=0.9\textwidth]
    {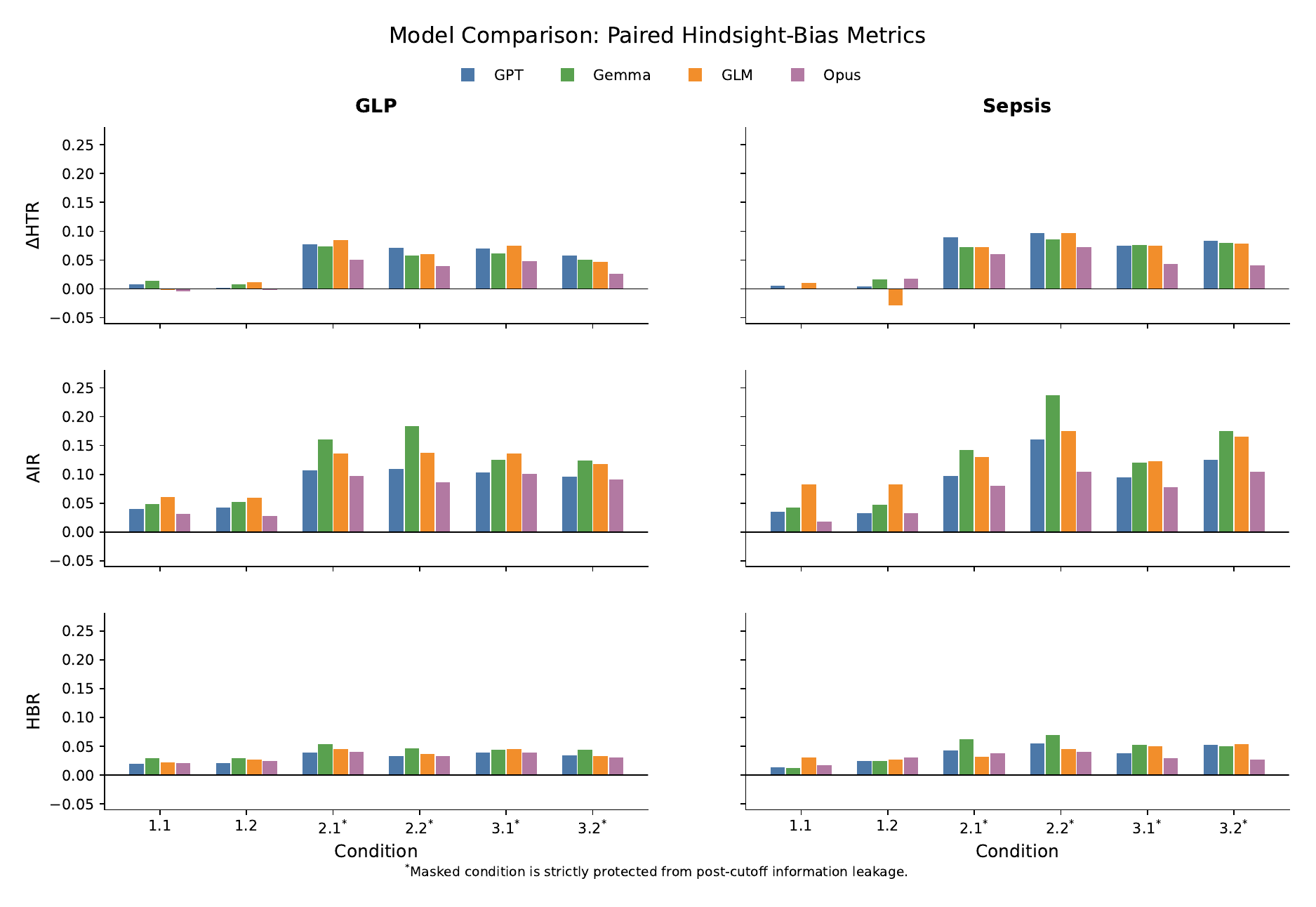}
    \caption{\textbf{Paired hindsight-bias metrics across models.}
    $\Delta$HTR, Answer Instability Rate (AIR), and Hindsight Bias Rate
    (HBR) for GPT, Gemma, GLM, and Opus on GLP and Sepsis. Asterisks
    denote strict masking conditions.}
    \label{fig:model_comparison_hindsight}
\end{figure*}

The paired metrics reinforce this pattern (Figure~\ref{fig:model_comparison_hindsight}). $\Delta$HTR was near zero in the 1.x conditions but consistently positive under strict masking, while AIR also increased substantially in the 2.x and 3.x conditions. Positive HBR further indicates that some answer changes were directional, shifting from the prospective reference under masking toward the hindsight trap when post-cutoff information was restored. Although effect sizes varied across models and datasets, the same qualitative pattern appeared across all four model families.

Cross-model differences should be interpreted cautiously. GPT was the only answerer from the same family as the question generator, and GPT was also used to judge free-response answers; however, the similar strict-masking pattern among Gemma, GLM, and Opus suggests that the overall effect is not specific to GPT. The plots also suggest that lower-performing model--condition combinations can exhibit greater degradation under masking, but this relationship is not monotonic across models or datasets. Thus, overall model quality may contribute to robustness, but accuracy and
susceptibility to hindsight information appear to capture distinct aspects of model behavior.

\clearpage

\section{Sensitivity Analysis with Explicit Hindsight Warning}
\label{app:modified_prompt}



\begin{figure}[htbp]
    \centering
    \includegraphics[width=\columnwidth]
        {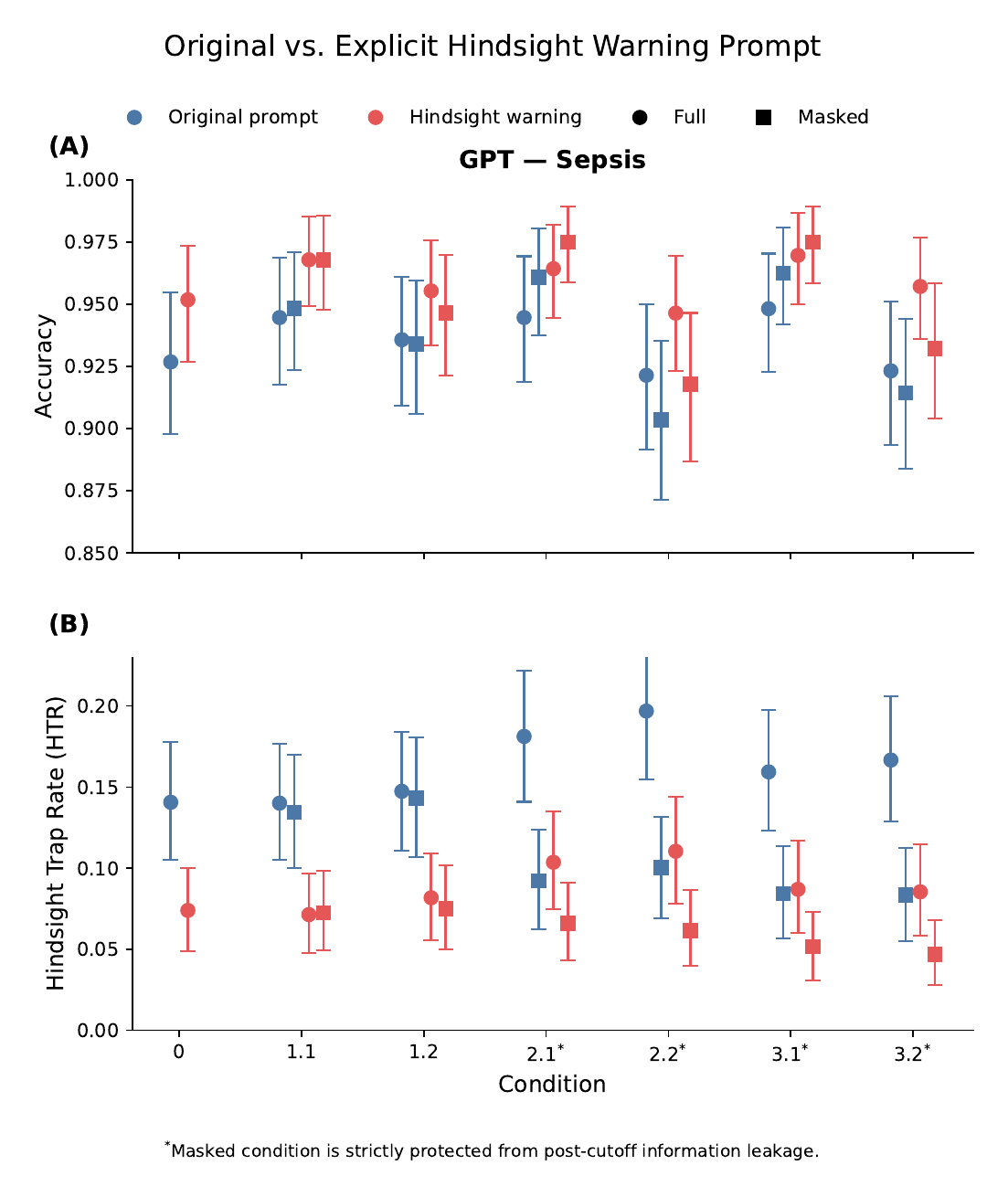}
    \caption{Accuracy and hindsight trap rate (HTR) across conditions
    for GPT under both prompt versions. Error bars show case-level 95\% bootstrap confidence
    intervals.}
    \label{fig:gpt_prompt_comparison_acc_htr}
\end{figure}
\begin{figure}[htbp]
    \centering
    \includegraphics[width=\columnwidth]
        {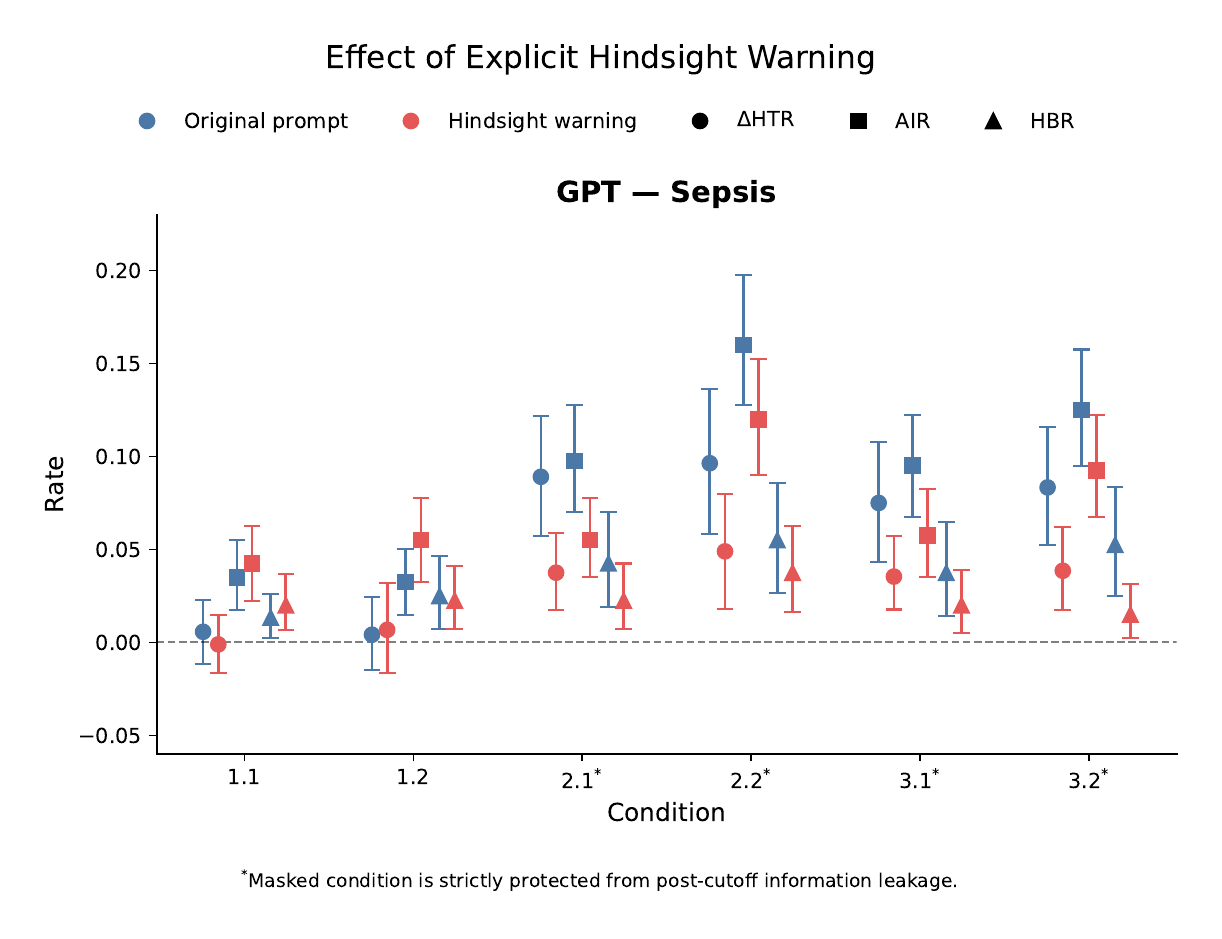}
    \caption{Hindsight bias metrics across paired conditions for GPT
    under both prompt versions.
    Metrics include the change in hindsight trap rate
    ($\Delta$HTR), AIR, and HBR. Error bars show case-level 95\%
    bootstrap confidence intervals.}
    \label{fig:gpt_prompt_comparison_paired}
\end{figure}
\begin{figure}[htbp]
    \centering
    \includegraphics[width=\columnwidth]
        {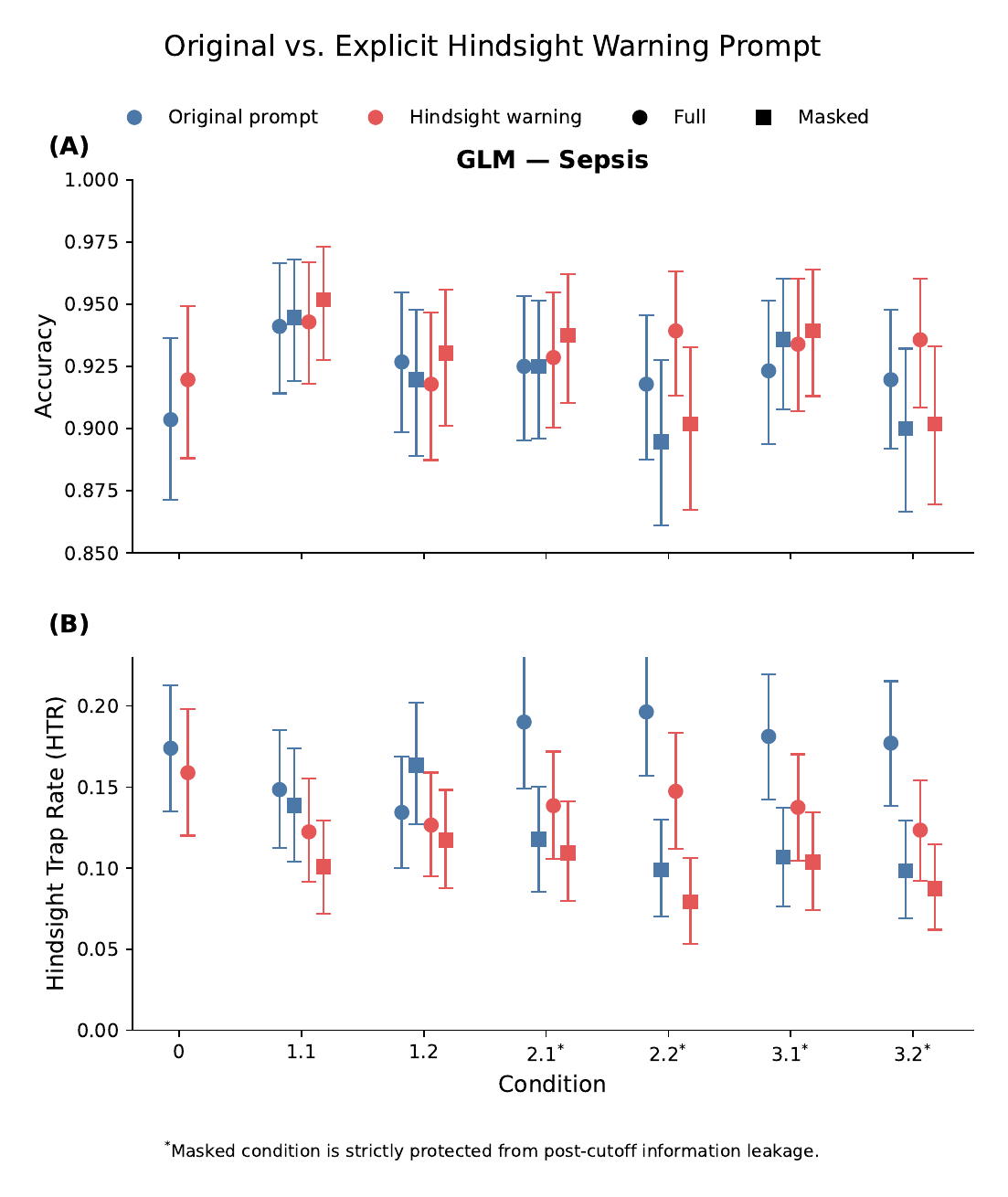}
    \caption{Accuracy and hindsight trap rate (HTR) across conditions
    for GLM under both prompt versions. Error bars show case-level 95\% bootstrap confidence
    intervals.}
    \label{fig:glm_prompt_comparison_acc_htr}
\end{figure}
\begin{figure}[htbp]
    \centering
    \includegraphics[width=\columnwidth]
        {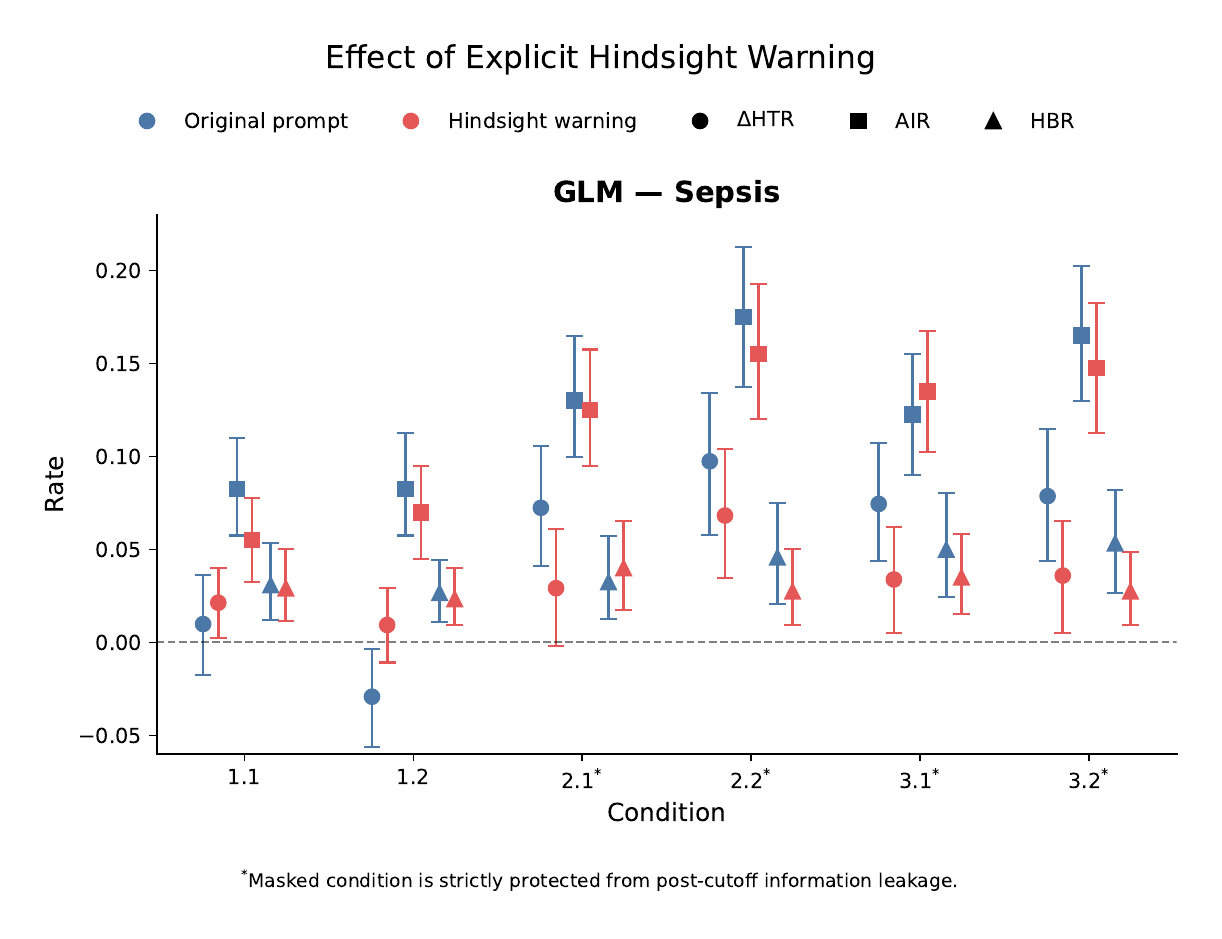}
    \caption{Hindsight bias metrics across paired conditions for GLM
    under both prompt versions.
    Metrics include the change in hindsight trap rate
    ($\Delta$HTR), AIR, and HBR. Error bars show case-level 95\%
    bootstrap confidence intervals.}
    \label{fig:glm_prompt_comparison_paired}
\end{figure}

Figures~\ref{fig:gpt_prompt_comparison_acc_htr}--\ref{fig:gpt_prompt_comparison_paired}
and~\ref{fig:glm_prompt_comparison_acc_htr}--\ref{fig:glm_prompt_comparison_paired}
report the sensitivity analysis using explicit hindsight warning
for GPT and GLM, respectively. Error bars indicate case-level 95\% bootstrap
confidence intervals. The modification in prompts can be found in Appendix~\ref{app:prompts}.

Across both
models, explicit hindsight warning attenuated hindsight-biased
responding but did not eliminate it. The reduction was most apparent in
Conditions~2.1--3.2, where full-condition HTR, $\Delta$HTR, and HBR were
generally lower under the modified prompt. AIR was also reduced,
particularly for GPT, but remained positive across the paired conditions,
as did HBR, indicating persistent hindsight-sensitive answer changes
despite the explicit warning. Accuracy remained broadly similar between
prompt versions for both models, suggesting that the reductions in
hindsight-related metrics were not driven by a general deterioration in
task performance. The consistency of this pattern across GLM and GPT
indicates that models can respond to explicit instructions against
hindsight bias, but that simple prompt-level mitigation is insufficient
to eliminate the effect.

\clearpage

\section{Human Validation}
\label{app:validation}
\input{appendix/human_validation}

\end{document}

%% file: appendix/lit_review_detailed.tex
\paragraph{General temporal reasoning}
Several benchmarks evaluate temporal reasoning in general domain LLMs. TRAM covers event order, arithmetic, frequency, and duration \citep{wang2024tram}; TimeBench provides a hierarchical evaluation across a broad range of temporal phenomena \citep{chu2024timebench}; Test of Time uses controlled synthetic tasks to study temporal logic and sensitivity to problem structure \citep{fatemi2025testoftime}; and TimE evaluates multi-level reasoning over Wikipedia, news, and dialogue scenarios \citep{wei2025time}. Most closely related, ExAnte evaluates inference under explicit temporal cutoffs and measures leakage from post-cutoff knowledge across several non-clinical tasks \citep{liu2026exante}. These benchmarks establish that temporal constraints remain difficult for LLMs, but they do not study clinical judgments or quantify whether exposing the later course of the same patient trajectory moves an answer toward a clinically defined hindsight trap.

\paragraph{Clinical question answering over longitudinal records.}
Clinical benchmarks increasingly test LLMs on realistic or longitudinal EHR tasks. MedAlign contains clinician-generated instructions grounded in longitudinal records \citep{fleming2024medalign}; MIMIC-Instr provides large-scale instruction-following examples derived from MIMIC-IV \citep{wu2024mimicinstr}; and EHRNoteQA evaluates questions spanning one or more discharge summaries \citep{kweon2024ehrnoteqa}. EHRSQL tests text-to-SQL reasoning over structured EHR databases, including time-sensitive questions and temporal expressions \citep{lee2022ehrsql}. TIMER explicitly links instructions to timestamps and evaluates temporal boundary adherence, trend detection, and chronological precision in longitudinal EHRs \citep{cui2025timer}. ASCENT evaluates stepwise diagnostic reasoning as evidence accumulates under incomplete information \citep{choi2026ascent}. RealICU evaluates windowed ICU decision support using prefix-only observations and labels produced after physicians review the full trajectory \citep{shen2026realicu}; unlike our paired design, it uses hindsight to construct reference labels rather than measuring how a model's own answer changes after outcome exposure. These works evaluate retrieval, temporal synthesis, instruction following, evolving diagnosis, or prospective performance against retrospectively informed labels. We instead hold the question and prospective reference target fixed and compare masked with complete-record exposure to measure the directional effect of later information on the model's response.

\paragraph{Reasoning from clinical case reports.}
Clinical case reports have also been used to construct challenging reasoning benchmarks. MedCaseReasoning pairs diagnostic cases with detailed reasoning statements derived from open-access case reports and evaluates both final diagnosis and reasoning recall \citep{wu2025medcasereasoning}. MedR-Bench structures case reports into examination recommendation, diagnostic decision-making, and treatment-planning tasks \citep{qiu2025medrbench}. Related work has reconstructed time-localized sepsis trajectories from PubMed Open Access case reports as textual time series \citep{noroozizadeh2026sepsistts}. These resources improve the realism and granularity of clinical reasoning evaluation, but primarily assess whether a model reaches or explains a final clinical conclusion. Our benchmark instead asks when a conclusion was supportable and measures how the same answer changes when post-cutoff events become visible.

%% file: appendix/experimental_conditions.tex
We evaluated seven input conditions (Table~\ref{tab:conditions}), each defining what information the answering model has access to when responding to a question. The question set itself was identical across all conditions: all questions were generated from the original narrative $N_i$ paired with the full human TTS $T_i^{\mathrm{H}}$, and the answering model never sees the generation inputs or the hidden evaluation metadata.\

\paragraph{Condition~0}
Condition~0 serves as a naive baseline representing the setting most common in current clinical QA systems, where the model is given a complete retrospective case report and asked to answer questions without any temporal masking. Because case reports are written retrospectively and may reference diagnoses, treatments, and outcomes in any order regardless of when they occurred, there is no reliable way to mask post-cutoff information from a narrative. Condition~0 therefore reflects the upper bound of potential hindsight exposure under narrative-only input.

\paragraph{Condition~1.x}
Conditions~1.x address this limitation partially by pairing the original narrative with a temporally masked TTS. The narrative component is held fixed and complete across the full and masked variants; only the TTS is truncated at the cutoff. While masking the TTS removes explicit post-cutoff temporal data, the original narrative may still implicitly contain outcome information, meaning that conditions~1.x achieve only partial information control.

\paragraph{Condition~2.x}
Conditions~2.x remove the narrative entirely, providing the TTS alone. This yields the cleanest paired comparison between prefix-only and complete temporal information, as the TTS is the sole information source and can be precisely truncated at $c_q$.

\paragraph{Condition~3.x}
Conditions~3.x also achieve rigorous temporal control. Conditions~3.x are motivated by the complementary limitations of conditions~1.x and~2.x: while condition~1.x retains a narrative but cannot fully mask it, and condition~2.x achieves strict masking but strips away the clinical context that a narrative provides, conditions~3.x combine both desiderata by supplying a narrative that is simultaneously present and strictly temporally controlled. The synthetic narrative presented to the answering model is derived from the same masked TTS used in that condition variant, as described in Section~\ref{sec:representations}. Because the synthetic narrative is generated from pre-cutoff input, it cannot contain implicit references to subsequent events. Paired with a masked TTS, conditions~3.x ensure that all information available to the answering model --- both narrative and temporal --- is strictly limited to what was observable before the cutoff, a guarantee that conditions~1.x cannot provide.

%% file: appendix/all_prompts.tex
\begin{minipage}[!htbp]{0.99\textwidth}
\begin{tcolorbox}[
  breakable,
  colback=teal!5,
  colframe=teal!70!black,
  title=Question generation prompt,
  boxrule=1pt,
  before skip=4pt,
  after skip=4pt,
]
\scriptsize
\setlength{\parskip}{1pt}

You are generating difficult clinical reasoning questions from case reports to build a benchmark evaluating accuracy and hindsight bias in language models. Questions should be genuinely hard---the answering model should make errors, especially by using outcome knowledge to revise past assessments it should not revise.

\vspace{5pt}
Generate 10 questions per case report with a mix of formats (free response, multiple choice with 4 options, true/false) and the following distribution by evaluation target:
\begin{itemize}
    \item \texttt{accuracy} (20\%): verifiable from case facts; answer does not change with more timeline information
    \item \texttt{hindsight\_bias} (30\%): answer is judgment-based (likelihood, severity, risk); no objective ground truth, but a model with outcome knowledge is strongly tempted to revise its answer
    \item \texttt{both} (50\%): defensible correct answer anchored to case facts or clinical guidelines, AND strongly tempting to revise with outcome knowledge
\end{itemize}

\vspace{5pt}
\textbf{Rules:}
\begin{itemize}
    \item Questions must be clinically focused, not meta-questions about the data. Never phrase a question in terms of what the TTS ``establishes,'' ``shows,'' ``records,'' or ``contains.'' Ask about the clinical facts themselves (e.g., ``Did angiography occur before embolization?'' not ``Does the TTS establish a strict chronological sequence between angiography and embolization?'').
    \item Do not include any time reference in the question text. Questions must be atemporal---e.g., ``What was the probability of a serious infectious complication?'' not ``At readmission, what was the probability...''. The answering model's timeline is controlled externally through masking; an explicit time reference tips off prospective reasoning.
    \item Anchor to early ambiguous presentations where findings were indistinguishable from benign alternatives, not to late confirmed events.
    \item The hindsight trap must be the most attractive answer for a model that knows the outcome---the correct prospective answer should feel conservative or uncertain by comparison.
    \item No more than 2 questions of the same reasoning type (likelihood, severity, causal attribution, treatment appropriateness, diagnostic classification, temporal ordering).
    \item For ordinal multiple choice questions: use exact probability ranges instead of vague labels (e.g., ``A. $<$10\%'', ``B. 10--30\%'', ``C. 30--60\%'', ``D. $>$60\%''). The correct answer and hindsight trap must be at opposite ends of the scale---if the correct answer is the lowest option, the hindsight trap must be the highest, and vice versa.
\end{itemize}

\vspace{5pt}
You are also provided with the textual time series (TTS) for this case---a JSON array of \texttt{\{"event": "...", "time": <hours>\}} tuples ordered chronologically by time.

\vspace{5pt}
\textbf{Use the TTS to:}
\begin{itemize}
    \item Identify events that the narrative describes out of chronological order. These are especially valuable for hindsight bias questions---the narrative's non-chronological framing may lead a model reading only the narrative to assume a different temporal order than what actually occurred. Design questions that target these moments, where a model relying on narrative framing would answer differently than one reasoning from the true chronological sequence.
    \item Identify tight event clusters (multiple events at similar timestamps) that represent moments of rapid clinical change---these are high-yield anchors for ambiguous early presentations.
\end{itemize}

\vspace{5pt}
Return a valid JSON array only. Each element:

{\ttfamily
\raggedright
\sloppy
\{"question": "...",\\
"format": "free response | multiple choice | true/false",\\
"options": ["A. ...", "B. ...", "C. ...", "D. ..."] or null,\\
"options\_type": "ordinal | categorical | null",\\
"eval\_target": "accuracy | hindsight\_bias | both",\\
"reference\_time": "clinical moment the question is implicitly about",\\
"early\_cutoff": \{\\
\hspace*{1em}"event": "specific event label from narrative just after reference time",\\
\hspace*{1em}"time": "time expression as written in the TTS"\\
\},\\
"correct\_answer": "...",\\
"hindsight\_trap": "the most tempting wrong answer for a model with\\
\hspace*{1em}full-case access; null if eval\_target is accuracy",\\
"hindsight\_trap\_explanation": "one sentence naming the post-cutoff fact\\
\hspace*{1em}that makes the trap so tempting; null if eval\_target is accuracy"\\
\}
}

\end{tcolorbox}
\end{minipage}

\begin{tcolorbox}[
  colback=teal!5,
  colframe=teal!70!black,
  title=Answer generation prompt: Condition 0 (Narrative only),
  boxrule=2pt,
]
\scriptsize
\setlength{\parskip}{1pt}

You answer clinical reasoning questions using only the provided case report.

\vspace{5pt}
For each request, you will receive:
\begin{enumerate}
    \item A case report.
    \item One JSON question object containing only these fields: \texttt{"question"}, \texttt{"format"}, and \texttt{"options"}.
\end{enumerate}

Answer only from the visible case report and the visible question object. Use general medical knowledge only to interpret clinical language, not to invent patient-specific facts.

\vspace{5pt}
\textbf{Rules:}
\begin{itemize}
    \item If \texttt{"format"} is \texttt{"true/false"}, answer exactly \texttt{"true"} or \texttt{"false"}.
    \item If \texttt{"format"} is \texttt{"multiple choice"}, answer with exactly one option string from \texttt{"options"}, including the option letter and text.
    \item If \texttt{"format"} is \texttt{"free response"}, answer concisely in one to three sentences.
    \item Always include an \texttt{"evidence"} field explaining why you gave the answer.
    \item The evidence should be a concise case-grounded rationale, ideally naming the relevant event, symptom, test, treatment, timing, or absence of needed information from the case report.
    \item Do not include reasoning, citations, markdown, or extra text outside the JSON.
\end{itemize}

\vspace{5pt}
Return only valid JSON in this exact shape:

\begin{verbatim}
{"answer": "...", "evidence": "..."}
\end{verbatim}

\end{tcolorbox}

\begin{tcolorbox}[
  colback=teal!5,
  colframe=teal!70!black,
  title=Answer generation prompt: Conditions 1 and 3 (Narrative + TTS),
  boxrule=2pt,
]
\scriptsize
\setlength{\parskip}{1pt}

You answer clinical reasoning questions using the provided case report and its textual time series (TTS).

\vspace{5pt}
For each request, you will receive:
\begin{enumerate}
    \item A case report.
    \item A TTS---a JSON array of \texttt{\{"event": "...", "time": <hours>\}} tuples ordered chronologically by time relative to an arbitrary anchor event.
    \item One JSON question object containing only these fields: \texttt{"question"}, \texttt{"format"}, and \texttt{"options"}.
\end{enumerate}

Answer using both the case report and the TTS. When they conflict, trust the TTS for event ordering and timing, and the narrative for clinical detail.

Use general medical knowledge only to interpret clinical language, not to invent patient-specific facts.

\vspace{5pt}
\textbf{Rules:}
\begin{itemize}
    \item If \texttt{"format"} is \texttt{"true/false"}, answer exactly \texttt{"true"} or \texttt{"false"}.
    \item If \texttt{"format"} is \texttt{"multiple choice"}, answer with exactly one option string from \texttt{"options"}, including the option letter and text.
    \item If \texttt{"format"} is \texttt{"free response"}, answer concisely in one to three sentences.
    \item Always include an \texttt{"evidence"} field explaining why you gave the answer.
    \item The evidence should be a concise rationale grounded in the case report or TTS, naming the relevant event, timestamp, symptom, test, treatment, or absence of needed information.
    \item Do not include reasoning, citations, markdown, or extra text outside the JSON.
\end{itemize}

\vspace{5pt}
Return only valid JSON in this exact shape:

\begin{verbatim}
{"answer": "...", "evidence": "..."}
\end{verbatim}

\end{tcolorbox}

\begin{tcolorbox}[
  colback=teal!5,
  colframe=teal!70!black,
  title=Answer generation prompt: Condition 2 (TTS only),
  boxrule=2pt,
]
\footnotesize
\setlength{\parskip}{1pt}

You answer clinical reasoning questions using only the provided textual time series (TTS).

\vspace{5pt}
For each request, you will receive:
\begin{enumerate}
    \item A TTS---a JSON array of \texttt{\{"event": "...", "time": <hours>\}} tuples ordered chronologically by time relative to an arbitrary anchor event.
    \item One JSON question object containing only these fields: \texttt{"question"}, \texttt{"format"}, and \texttt{"options"}.
\end{enumerate}

Answer using only the TTS. Use general medical knowledge only to interpret clinical language, not to invent patient-specific facts.

\vspace{5pt}
\textbf{Rules:}
\begin{itemize}
    \item If \texttt{"format"} is \texttt{"true/false"}, answer exactly \texttt{"true"} or \texttt{"false"}.
    \item If \texttt{"format"} is \texttt{"multiple choice"}, answer with exactly one option string from \texttt{"options"}, including the option letter and text.
    \item If \texttt{"format"} is \texttt{"free response"}, answer concisely in one to three sentences.
    \item Always include an \texttt{"evidence"} field explaining why you gave the answer.
    \item The evidence should be a concise rationale grounded in the TTS, naming the relevant event, timestamp, or absence of needed information.
    \item Do not include reasoning, citations, markdown, or extra text outside the JSON.
\end{itemize}

\vspace{5pt}
Return only valid JSON in this exact shape:

\begin{verbatim}
{"answer": "...", "evidence": "..."}
\end{verbatim}

\end{tcolorbox}

\begin{tcolorbox}[
  colback=teal!5,
  colframe=teal!70!black,
  title=Hindsight warning prompt,
  boxrule=2pt,
]
\footnotesize
\setlength{\parskip}{1pt}

For the sensitivity analysis experiments, the answer-generation prompts for Conditions 0--3 were identical to the corresponding prompts above, except that the following two sentences were appended to the beginning (for GPT) or end (for GLM) of each prompt:

\vspace{5pt}

\textit{Use your medical expertise for this reasoning task; do not let hindsight cloud your judgment. Use only knowledge available at the time; do not let the final outcome of the case affect your answer.}

\end{tcolorbox}

\begin{tcolorbox}[
  breakable,
  colback=teal!5,
  colframe=teal!70!black,
  title=Synthetic case generation prompt: Condition 3,
  boxrule=1pt,
  before skip=4pt,
  after skip=4pt,
]
\scriptsize
\setlength{\parskip}{1pt}

You are a medical expert tasked with reconstructing a synthetic clinical case report from a patient timeline presented in textual time-series (TTS) format. The input contains two columns: event and time in hours, where the timeline conventions are as follows: Time 0 represents the patient's hospital admission; events occurring before admission have negative time values and events occurring after admission have positive time values.

\vspace{5pt}
\textbf{Guidelines:}
\begin{itemize}
    \item Use clinical and natural language to convert fragmented timeline entries into a readable narrative.
    \item Follow the order in which events appear in the TTS file as the narrative sequence.
    \item Central events include: symptom onset or major clinical changes; hospital admission, transfer, discharge, or death; diagnostic tests and clinically significant results; diagnoses; procedures and interventions; medication initiation, discontinuation, or major adjustment; complications and treatment responses.
    \item Express the timing of central events relative to nearby central events (e.g., ``two weeks after that,'' ``one month after presentation,'' ``shortly after admission'') using any unit of time (hours, days, weeks, months, years) that is the most natural to read.
    \item Use a standalone or absolute timepoint only when there is no clear anchoring event or when it is clearer than a relative phrase.
    \item Include only information explicitly supported by the input timeline.
    \item Do not infer, invent, or add diagnoses, symptoms, treatments, results, outcomes, demographic details, or causal relationships that are not documented.
    \item When the timeline is ambiguous or incomplete, report the information conservatively without any speculation.
\end{itemize}

\vspace{5pt}
\textbf{Output Requirements:} Return only the synthetic clinical case report. Do not include headings about the task, explanations, notes, disclaimers, or commentary outside the report.

\vspace{3pt}
\textbf{Example input:}

{\ttfamily\scriptsize
42-year-old | 0\\
female | 0\\
no pathological history | 0\\
consulted the rheumatology clinic | 0\\
initiated subcutaneous semaglutide for weight loss | -1008\\
starts complaining of myalgia | -672\\
edema in the four limbs | -672\\
review of symptoms was negative | -672\\
new contrast MRI took place | +730\\
notable decrease in inflammatory changes in the limb muscles | +730
}

\vspace{3pt}
\textbf{Example output:}

A 42-year-old female patient with no pathological history consulted
the rheumatology clinic. Six weeks before the consultation, she had
initiated subcutaneous semaglutide for weight loss. Two weeks after
that, she starts complaining of myalgia and edema in the four limbs.
The review of symptoms by systems was negative. One month after
presentation, a new contrast MRI showed a notable decrease in the
extent of inflammatory changes in the limb muscles.

\end{tcolorbox}

\begin{tcolorbox}[
  colback=teal!5,
  colframe=teal!70!black,
  title=GPT judge prompt: Individual metrics for free-response questions,
  boxrule=2pt,
]
\footnotesize
\setlength{\parskip}{1pt}

You are evaluating an answer to a clinical temporal-QA question.

Return only JSON. Decide clinical meaning, not exact wording.

\begin{itemize}
    \item \texttt{accuracy\_score} must be 1 for fully correct, 0.5 for partially correct, and 0 for incorrect.
    \item \texttt{htr\_score} must be 1 if the answer fully matches the \texttt{hindsight\_trap}, 0.5 if it partially reflects the hindsight trap, and 0 if it does not reflect the trap.
    \item If \texttt{hindsight\_trap} is null, \texttt{htr\_score} must be 0.
\end{itemize}

\end{tcolorbox}

\begin{tcolorbox}[
  colback=teal!5,
  colframe=teal!70!black,
  title=GPT judge prompt: Paired metrics for free-response questions,
  boxrule=2pt,
]
\footnotesize
\setlength{\parskip}{1pt}

You are evaluating paired answers to a clinical temporal-QA question.

Return only JSON. Decide clinical meaning, not exact wording.

\begin{itemize}
    \item \texttt{full\_equals\_masked} is true if the full-condition and masked-condition answers mean the same thing clinically.
    \item For \texttt{full\_accuracy\_score} and \texttt{masked\_accuracy\_score}, use 1 for fully correct, 0.5 for partially correct, and 0 for incorrect.
    \item For \texttt{full\_htr\_score} and \texttt{masked\_htr\_score}, use 1 if the answer fully matches the \texttt{hindsight\_trap}, 0.5 if it partially reflects the hindsight trap, and 0 if it does not reflect the trap.
    \item If \texttt{hindsight\_trap} is null, both HTR scores and \texttt{hbr\_score} must be 0.
    \item For \texttt{hbr\_score}, give hindsight-bias credit only to the extent that the masked answer is correct and the full answer reflects the trap; use 1, 0.5, or 0.
\end{itemize}

\end{tcolorbox}
\twocolumn

%% file: appendix/masking_cutoff.tex
To characterize where masking occurs within the clinical timelines, we
define the relative masking cutoff position as the chronological index of
the cutoff event within each TTS, normalized to the interval $[0,1]$,
where 0 and 1 correspond to the beginning and end of the timeline,
respectively. This measure reflects position within the sequence of
recorded events rather than elapsed clinical time.

Table~\ref{tab:cutoff_distribution} summarizes the distribution of relative cutoff positions, $\operatorname{RelPos}(q)$, across the two datasets. In both cases, the mean and median were near the midpoint of the timeline, and the standard deviations indicate substantial spread across temporal positions. The observed ranges were also broad, showing that questions span much of the clinical trajectory rather than clustering within a narrow temporal window. 

\begin{table}[!h]
\centering
\footnotesize
\caption{Distribution of relative cutoff position $\operatorname{RelPos}(q)$ across questions in each dataset.}
\label{tab:cutoff_distribution}
\begin{tabular}{lcc}
\toprule
& \textbf{Sepsis} & \textbf{GLP} \\
\midrule
Mean     & 0.499 & 0.565   \\
Median   & 0.500 & 0.581   \\
SD       & 0.206 & 0.206   \\
Min      & 0.029 & 0.000   \\
Max      & 0.945 & 1.000   \\
\bottomrule
\end{tabular}
\end{table}

\begin{figure}[!htbp]
    \centering
    \includegraphics[width=\columnwidth]{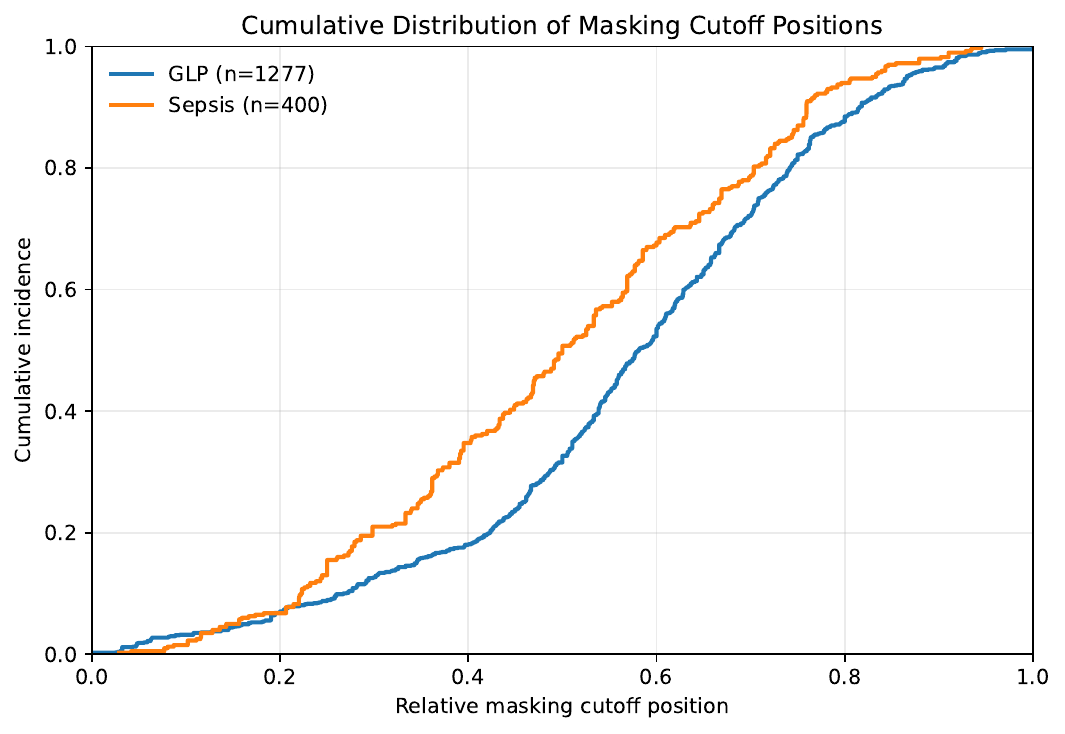}
    \caption{Empirical cumulative distribution of relative masking cutoff
    positions for GLP and Sepsis. Relative position is defined using the
    chronological index of the cutoff event within the TTS and normalized
    to $[0,1]$, with 0 representing the beginning and 1 the end of the
    recorded event sequence.}
    \label{fig:masking_cutoff_distribution}
\end{figure}

Figure~\ref{fig:masking_cutoff_distribution} shows that masking points are
distributed broadly across the clinical timeline rather than being
concentrated at a single position. Sepsis cutoffs tend to occur somewhat
earlier than GLP cutoffs, particularly in the middle of the timeline,
although both datasets span nearly the full range of relative positions.
This distribution indicates that the prospective evaluation is conducted
at diverse stages of the underlying clinical trajectories rather than at
a fixed or narrowly defined point in each case.

%% file: appendix/human_validation.tex
This appendix provides additional details on the human validation
described in Section~\ref{sec:human_validation}, including the
error-enriched sampling procedure and validation results.

\paragraph{Annotators and review procedure.}
Four clinical annotators participated: one physician and three medical
or physician assistant trainees. We sampled 10\% of cases for review
and divided them into two non-overlapping 5\% subsets, each independently
evaluated by two annotators. For each question, annotators assessed
(1) whether the question was clinically reasonable, (2) whether they
agreed with the prospective reference answer, and (3) whether they
agreed with the hindsight-trap answer. All judgments were binary
(yes/no).

\paragraph{Error-enriched sampling.}
Rather than sampling uniformly, we preferentially selected cases in
which the evaluated model exhibited errors or hindsight-sensitive
behavior. For each case, we computed an error-proneness score (EPS)
from four model-derived quantities: $1-\mathrm{Accuracy}$ and HTR under
Condition~0, and AIR and HBR under Condition~2.1. Each quantity was
aggregated to the case level and min--max normalized within cohort:

\begin{equation}
\begin{aligned}
\mathrm{EPS}_i ={}&
0.4\,\widetilde{(1-\mathrm{Accuracy})}_i
+ 0.2\,\widetilde{\mathrm{HTR}}_i \\
&+ 0.2\,\widetilde{\mathrm{AIR}}_i
+ 0.2\,\widetilde{\mathrm{HBR}}_i .
\end{aligned}
\end{equation}

Cases were sampled without replacement with weights
\[
w_i \propto \exp\left(\frac{\mathrm{EPS}_i}{\tau}\right),
\qquad \tau=0.3.
\]
Thus, a 0.3-unit increase in EPS corresponds to an approximately
$e\approx2.72$-fold increase in relative sampling weight. This design
provides a targeted stress test of benchmark annotations in cases most
relevant to the observed model errors and hindsight effects.

\paragraph{Human validation results.}
The validation sample comprised 170 questions, each reviewed by two
annotators. Agreement with the benchmark annotations was high across
question reasonableness, prospective reference answers, and
hindsight-trap answers, with sample agreement ranging from 95.9\% to
98.7\% (Table~\ref{tab:human_validation_weighted}). Inter-annotator
agreement was similarly high (91.8--99.2\%).

Because sampling favored cases with higher EPS, we additionally report
inverse-probability-weighted estimates based on the cohort-specific
sampling design. These closely matched the corresponding sample
estimates (Table~\ref{tab:human_validation_weighted}) and should be
interpreted as design-weighted point estimates; uncertainty intervals
would require case-level resampling.

\begin{table*}[!t]
\centering
\caption{Human validation results. Sample estimates are calculated from
the error-enriched validation set; weighted estimates use
inverse-probability weights from the cohort-specific sampling design.}
\label{tab:human_validation_weighted}

\small
\setlength{\tabcolsep}{12pt}
\renewcommand{\arraystretch}{1.12}

\begin{tabular}{@{}lccc@{}}
\toprule
\textbf{Assessment}
& \textbf{Sample agreement}
& \textbf{Applicable assessments}
& \textbf{Weighted agreement} \\
\midrule

\multicolumn{4}{@{}l}{\textit{Agreement with benchmark annotations}} \\[2pt]
Question clinically reasonable
    & 95.9\% & 326/340 & 95.8\% \\
Prospective reference answer
    & 98.7\% & 235/238 & 99.0\% \\
Hindsight-trap answer
    & 97.1\% & 264/272 & 96.5\% \\

\addlinespace[4pt]
\multicolumn{4}{@{}l}{\textit{Inter-annotator agreement}} \\[2pt]
Question clinically reasonable
    & 91.8\% & 156/170 & 91.6\% \\
Prospective reference answer
    & 99.2\% & 118/119 & 99.5\% \\
Hindsight-trap answer
    & 94.1\% & 128/136 & 93.0\% \\

\bottomrule
\end{tabular}

\vspace{3pt}
\begin{minipage}{0.94\textwidth}
\footnotesize
\textit{Note.} Sample agreement is calculated from individual
annotator assessments. Inter-annotator agreement is the proportion of
applicable questions for which both annotators assigned to the subset
gave the same judgment. Not-applicable answer assessments are excluded
from the corresponding denominators.
\end{minipage}
\end{table*}